\documentclass[10pt,twocolumn,letterpaper]{article}

\usepackage[pagenumbers]{cvpr} % To force page numbers, e.g. for an arXiv version

\usepackage[dvipsnames]{xcolor}

\usepackage{amsmath}
\definecolor{cvprblue}{rgb}{0.21,0.49,0.74}
\definecolor{lightgreen}{RGB}{156, 235, 173}
\definecolor{lightblue}{RGB}{220, 239, 255}
\usepackage[pagebackref,breaklinks,colorlinks,citecolor=cvprblue]{hyperref}
\usepackage{xcolor,colortbl}
\usepackage{multirow}
\usepackage{array,booktabs}
\usepackage{fixltx2e}
\usepackage{amsmath}
\usepackage[normalem]{ulem}
\usepackage{hwemoji}
\usepackage{wasysym}
\usepackage{sidecap}
\usepackage{indentfirst}

\definecolor{lightgray}{HTML}{f2f2f2}

\newcommand{\cellcolorlightgray}{\cellcolor{lightgray}}

\def\paperID{20} % *** What's the Paper ID for workshop? 
\def\confName{CVPR}
\def\confYear{2024}

\title{OmniControlNet: Dual-stage Integration for Conditional Image Generation}

\author{%
  Yilin Wang$^{*,1}$ \qquad Haiyang Xu$^{*,2,4}$ \qquad Xiang Zhang$^{4}$ \qquad Zeyuan Chen$^{4}$ \\
  Zhizhou Sha$^{1}$ \qquad Zirui Wang$^{3}$ \qquad
  Zhuowen Tu$^{4}$ \\
  $^1$ Tsinghua University \quad $^2$ University of Science and Technology of China \\
  $^3$ Princeton University \quad $^4$ University of California, San Diego \\
}

\begin{document}

\twocolumn[{
\renewcommand\twocolumn[1][]{#1}
\maketitle
\begin{center}
    \centering
    \vspace*{-1.5em}
    \includegraphics[width=1.0\textwidth]{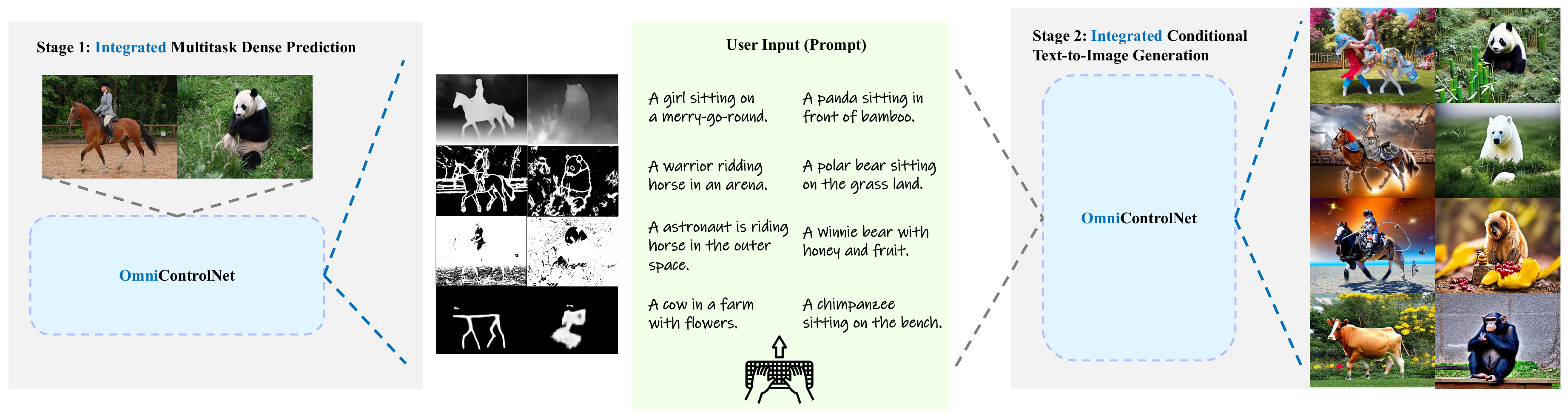}
    \captionsetup[figure]{hypcap=false}
    \captionof{figure}{\textbf{Given an input image, our single, integrated OmniControlNet} extracts its control features and generates high-quality images. From the first to the last row in the middle, the feature visualization represents Depth, HED, Scribble, and Animal Pose respectively.
    }
\label{fig:teaser}
\end{center}
}]

\begin{abstract}

\vspace{-0.5em}

We provide a two-way integration for the widely adopted ControlNet by integrating external condition generation algorithms into a single dense prediction method and incorporating its individually trained image generation processes into a single model. Despite its tremendous success, the ControlNet of a two-stage pipeline bears limitations in being not self-contained (e.g. calls the external condition generation algorithms) with a large model redundancy (separately trained models for different types of conditioning inputs). Our proposed OmniControlNet consolidates 1) the condition generation (e.g., HED edges, depth maps, user scribble, and animal pose) by a single multi-tasking dense prediction algorithm under the task embedding guidance and 2) the image generation process for different conditioning types under the textual embedding guidance. OmniControlNet achieves significantly reduced model complexity and redundancy while capable of producing images of comparable quality for conditioned text-to-image generation.
\let\thefootnote\relax\footnotetext{* equal contribution. Work done during the internship of Yilin Wang, Haiyang Xu, Zhizhou Sha, and Zirui Wang at UC San Diego.}

\vspace{-1.5em}

\end{abstract}

\section{Introduction}
\label{sec:introduction}

\begin{figure*}[h!]
  \centering
    \includegraphics[width=1\textwidth]{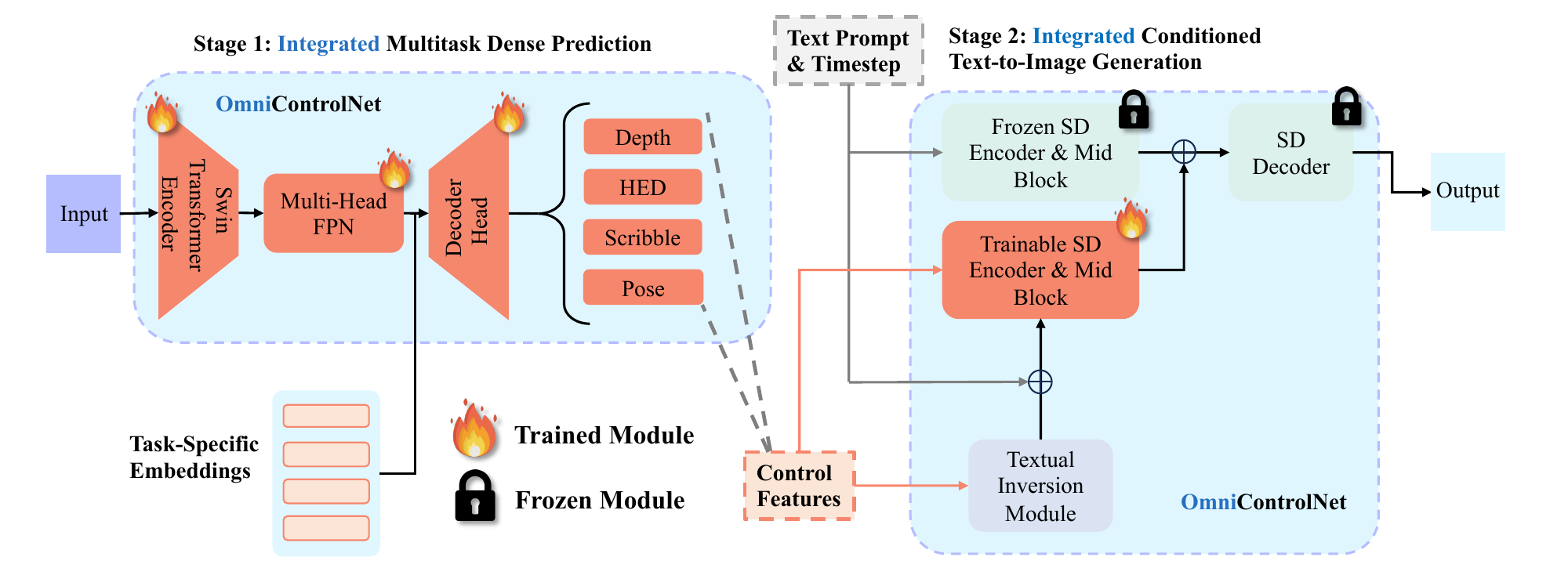}
    \vspace*{-1em}
    \caption{\textbf{Our OmniControlNet model.} From condition generation to image synthesis, while the ControlNet model has to deal with all the features separately, our model can handle the tasks within an integrated pipeline.}
    \vspace*{-1em}
    \label{fig:wholepipeline}
\end{figure*}

The exploding development of diffusion \cite{sohl2015deep,ho2020denoising,song2020denoising} based text-to-image generators \cite{ramesh2021zero,saharia2022photorealistic,rombach2022high,poole2022dreamfusion,ruiz2023dreambooth} has led to a recent wave of generative model progressing beyond traditional models such as VAE \cite{kingma2013auto} and GAN \cite{tu2007learning, goodfellow2014generative}.

The ControlNet \cite{zhang2023adding} further promotes the popularity of text-to-image generation by introducing additional user controls as the conditioning input available in a myriad of forms including edges \cite{4767851canny,xie15hed}, line segments \cite{gu2022towards}, human pose \cite{8765346openpose}, normal map \cite{vasiljevic2019diode}, depth map \cite{ranftl2020robust}, segmentation map \cite{zhou2019semantic}, and user scribble. With the additional image-level input beyond the text prompts, ControlNet can greatly expand the scope of application domains for text-to-image generation to real-world workflows in various areas, including design, architecture, gaming, art, manufacturing, animation, and human-computer interaction.

\begin{figure}[!htp]
    \centering
    \includegraphics[width=\linewidth,trim=20pt 0 20pt 0,clip]{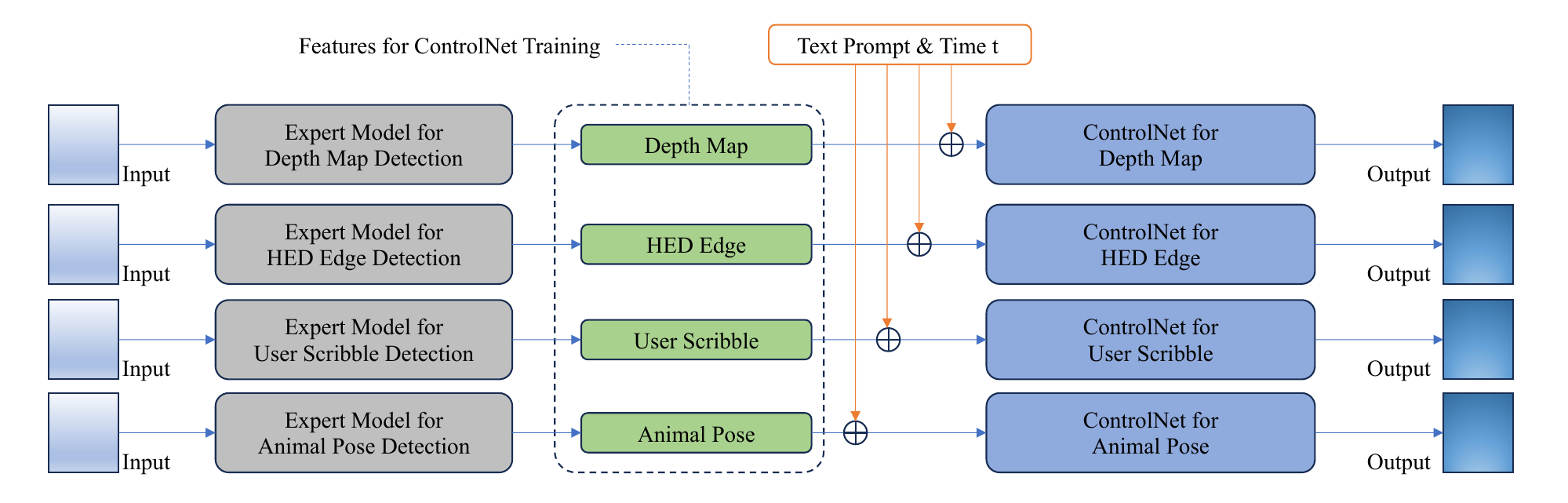}
  \footnotesize{$^*$ ControlNet for \textless\textit{feature}\textgreater is similar to our OmniControlNet's stage 2 in \cref{fig:wholepipeline}, except that there's no input from the textual embedding module.}
      \caption{\textbf{Original ControlNet \cite{zhang2023adding} model.} For different features, we have to use different expert models for condition generation, and we have to train ControlNet on each of the features.}
      \label{fig:ctnpipeline}
  \label{fig:pipeline_controlnet}
  % \vspace{-1em}
\end{figure}

ControlNet \cite{zhang2023adding} is a two-stage pipeline comprising 1) a condition generation stage and 2) a text-to-image generation stage conditioned on the output from the first stage. Despite the great success ControlNet has achieved, it still suffers from the issue of large model redundancy in two means: 1) in stage 1, a specific external algorithm is executed to create each type of image-level condition, and 2) in stage 2, a separate diffusion model is trained for each type of conditional input. \cref{fig:pipeline_controlnet} gives an schematic illustration for the ControlNet method \cite{zhang2023adding}.

In this paper, we aim to alleviate the algorithm and model redundancy problem in ControlNet \cite{zhang2023adding} by proposing OmniControlNet, which provides a dual-stage integration. That is, in stage 1, instead of calling the external algorithms, we develop an integrated dense image prediction method to perform edge detection, depth map generation, animal pose estimation, and scribble generation in a single multi-tasking framework under the guidance of task prompts; in stage 2, instead of training separate image generation models for different conditioning input types, we train a single model for four kinds of image-level conditional control under the textual inversion guidance. We observe a large model, parameter, and memory redundancy reduction, compared with the existing approaches, while being able to generate comparable image quality. The contribution of our work can be summarized as follows.

\vspace{0.2em}
The contribution of our paper is summarized as follows:

\begin{itemize}
    \item We develop a new module to integrate four dense image prediction tasks, including edge detection, depth estimation, scribble segmentation, and animal pose estimation, under the task embedding guidance.
    \item We develop a new module to perform conditioned text-to-image generation that integrates four different types of conditional input under the textual inversion guidance.
    \item Combining the above two modules yields OmniControlNet, which greatly reduces algorithm complexity for conditional text-to-image generation. OmniControlNet points to a promising direction for condition text-to-image generation under an integrated pipeline.
\end{itemize}

\vspace{-0.5em}
\section{Related Works}
\label{sec:related_work}

\subsection{Text-to-Image Generation}

The task of text-to-image generation \cite{ramesh2021zero,ding2021cogview,lin2021m6,yu2022scaling} is to generate an image matching the provided text prompts using deep learning models. Before the wide use of diffusion models, the task was primarily achieved by GAN \cite{goodfellow2014generative} based models  \cite{reed2016generative,zhang2017stackgan,xu2018attngan}. The work \emph{Generative Adversarial Text to Image Synthesis} \cite{reed2016generative} applied an encoder to encode the texts and concatenated the encoded features to the image features before inserting them into the GAN model, which was among the first works to tackle the task. After the introduction of diffusion models \cite{ho2020denoising,song2020denoising}, lots of diffusion-based models appeared \cite{chen2023anydoor,hao2023vico,balaji2022ediffi,meng2021sdedit,guo2023animatediff,feng2022training,hertz2022prompt,voynov2023sketch,singer2022make,wang2022pretraining,liu2022compositional,brooks2023instructpix2pix,tumanyan2023plug,huang2023region}, which mainly used cross attention to combine the image and text features in the UNet \cite{ronneberger2015u} backbone. DALLE-2 \cite{ramesh2022hierarchical} and Stable Diffusion \cite{rombach2022high} are among the outstanding literature in the field. Many works, including T2I-Adapter~\cite{mou2023t2i}, ControlNet \cite{zhang2023adding} our OmniControlNet model, are based on the Stable Diffusion model.

\vspace{-0.5em}
\subsection{Image-to-Image Generative Model}

Image-to-image generation involves transferring an image from one domain to another. For example, in ControlNet \cite{zhang2023adding}, additional features provided as images are fed into the model to generate the required images. Before the widespread use of diffusion models, GAN-based models \cite{goodfellow2014generative} such as \cite{alaluf2021only,choi2018stargan,gal2021stylegan,karras2019style,katzir2022multi,richardson2021encoding,park2019semantic,patashnik2021styleclip,wang2018high,zhu2017unpaired,zhu2017toward} and Transformer-based models \cite{vaswani2017attention,ramesh2021zero,esser2021taming} were commonly adopted. CycleGAN \cite{zhu2017unpaired} was one of the foremost models for image-to-image transfer, utilizing a GAN-based approach for style transfer with cycle consistency. With the introduction of diffusion models \cite{ho2020denoising,song2020denoising}, many \cite{chen2021pre,su2022dual,saharia2022palette} have demonstrated the significant potential of diffusion models in this task. Recently, several works \cite{zhang2023adding,li2023gligen,zhao2023uni,qin2023unicontrol,mou2023t2i,huang2023composer} have combined text and image conditions within diffusion models, enabling the generation of high-quality images. ControlNet is a notable example, taking text prompts and additional features as constraints to guide image generation.

\subsection{Condition Generation}

ControlNet has demonstrated its performance in conditional image generation across various conditions, including Depth Map \cite{ranftl2020robust}, Canny Edge \cite{xie15hed, 4767851canny}, OpenPose \cite{8765346openpose}, Normal Map \cite{vasiljevic2019diode}, User Scribble, and Segmentation \cite{park2019semantic}, \etc. In this section, we delve into four representative tasks: Depth Map, HED Edge, User Scribble, and Animal Pose, along with the expert models associated with each.

Generating depth maps to represent relative distances is a fundamental challenge in computer vision and 3D scene understanding tasks. Numerous methods have been proposed, ranging from traditional stereo matching algorithms \cite{4531745Make3D, 6909413pull} to deep learning-based approaches \cite{FuCVPR18-DORN, laina2016deeper, liu2015learning, xu2018structured}. We use MiDAS \cite{birkl2023midas, Ranftl2022midas} as our expert model, which exhibits exceptional performance and generalization capabilities.

Image edge detection plays a major role in tasks such as object segmentation and visual salience. Early methods \cite{Kittler1983sobel, marr1980theory, 1159946statedge, Pb, gPb} relied on manual design for edge detection. However, with the advent of deep learning, learning-based methods \cite{huan2021unmixing, wang2017deep, su2021pixel, 7299024deepcontour, Pu_2022_CVPREDTER} have demonstrated great potential in handling edge detection tasks. A classic benchmark in this field is Holistically-Nested Edge Detection (HED) \cite{xie15hed}, and we take it as our expert model.

User scribbles serve as user-defined guidance for image generation tasks, enabling users to convey their intentions and preferences to the generative model. In ControlNet, this involves a simple mapping of pixels with values greater than 127 to 255, and the rest to 0 in an image.

Generating pose maps, which encode spatial information about the arrangement of objects or characters in images, is crucial for tasks like image-to-image translation, particularly in human or object pose manipulation. Human pose estimation models \cite{newell2016hourglass, xiao2018simple, sun2019deep, cao2017realtime, yang2017learning} are designed to describe human skeletons. Notable benchmarks in this domain include PoseNet \cite{kendall2015posenet} and OpenPose \cite{8765346openpose}. In terms of animal pose estimation, which often presents more diversity and challenges than human pose estimation, datasets like AP-10K \cite{yu2021ap} and APT-36K \cite{yang2022aptk} are considered mainstream references.

\section{Background}
\label{sec:background}

\subsection{ControlNet}

The ControlNet model \cite{zhang2023adding} presents an efficient framework for fine-tuning the Stable Diffusion model \cite{rombach2022high}. It introduces an additional control feature (\eg depth map or edge detection) to the generative process, ensuring that the generated images adhere to both the textual prompt and the control condition. In our approach, the weights of the Stable Diffusion model (SD-v1.5) are fixed, while a trainable duplicate of the weights from the 12-layer U-Net encoder and middle block is created. The additional features are integrated into this trainable duplicate via a zero-convolution layer (a $1\times 1$ convolution layer with all-zero initial weights).

We denote the encoder in the frozen part as $\mathcal{E}$, the encoder of the trainable copy as $\mathcal{E}'$, the middle block and the decoder of the frozen part as $\mathcal{M}$ and $\mathcal{D}$, respectively. Let the CLIP-encoded additional feature be $c_f$, the input of the model as $z$, time as $t$, and the CLIP-encoded text prompt as $c_t$. With $\mathcal{Z}_1, \mathcal{Z}_2$ representing two trainable zero convolution layers, the output of the trainable copy should be $\mathcal{E'}(\mathcal{Z}_1(c_f)+z, t, c_t)$. Consequently, the output of the model, $\epsilon_{pred}$, which also estimates the noise in the denoising process, should be
\begin{equation}
    \epsilon_{pred} = \mathcal{D}(\mathcal{M}(\mathcal{E}(z, t, c_t) + \mathcal{Z}_2(\mathcal{E}'(\mathcal{Z}_1(c_f)+z, t, c_t))))
\end{equation}
During training, suppose the noise of a diffusion step be $\epsilon$, then the training loss should be 
\begin{equation}
    \mathcal{L}_{\text{diff}} = \Vert \epsilon - \epsilon_{\text{pred}} \Vert_{2}^{2}
\end{equation}

\section{Our Method}
\label{sec:method}

\subsection{Stage 1: Multi-task Dense Image Prediction}

\begin{figure}[h!]
  \centering
    \vspace{-1em}
    \includegraphics[width=\linewidth]{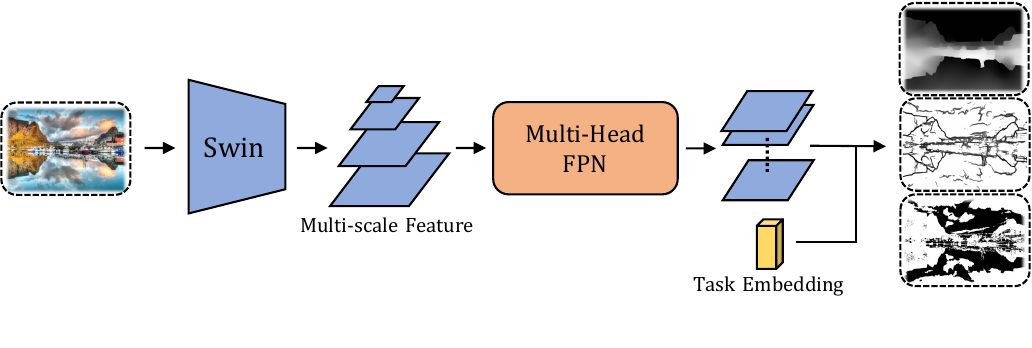}
    \vspace{-1em}
    \caption{\textbf{An overview of our multi-task dense image prediction pipeline.} First, we leverage a Swin Transformer to extract multi-scale features and propose a multi-head FPN to get full-resolution feature maps. Finally, we utilize task-specific embeddings to decode dense predictions from the feature maps. }
    \vspace{-0.5em}
    \label{fig:stage1}
\end{figure}

As depicted in \cref{fig:stage1}, our multi-task dense image prediction model is architecturally divided into three components: a backbone structure, a Multi-Head Feature Pyramid Network (FPN) \cite{lin2017fpn}, and a Decoder Head.

Initially, we employ a pre-trained Swin Transformer \cite{liu2021swin} to extract multi-scale image features. Considering the resolution of the input image as 1$\times$, the extracted features at each stage correspond to resolutions of $\frac{1}{4}\times$, $\frac{1}{8}\times$, $\frac{1}{16}\times$, and $\frac{1}{32}\times$, with a uniform feature channel count of 256.

Subsequently, a Multi-Head FPN is employed to harness rich semantic information from these multi-scale features. To foster feature diversity across various task types, the FPN is structured in a parallel configuration with $\mathbf{m}$ distinct heads, each representing a variant of the original FPN architecture. Specifically, each FPN head undergoes an additional transposed convolution layer to upscale the resolution to 1$\times$ while simultaneously reducing the channel dimension to $\mathbf{C}$. The concatenated outputs of all $\mathbf{m}$ heads yield a comprehensive, full-resolution multi-task output feature with channel dimension $\mathbf{mC}$.

In the final stage, task-specific embedding is leveraged to decode the target condition from the aforementioned output. The flexibility in the type of task embedding is noteworthy; both one-hot and clip text embeddings derived from the task name are effective. We employ a Multilayer Perceptron (MLP) to project the task embedding into a latent space with an embedding dimension of $\mathbf{mC}$, subsequently unsqueezing the channel dimension to 1. A cross-product operation is then executed between the output of the Multi-Head FPN and the encoded task embedding, culminating in the decoder output, followed by a Sigmoid.

\subsection{Stage 2: Conditioned T2I Generation}

\begin{figure*}[h!]
  \centering
    \includegraphics[width=\linewidth]{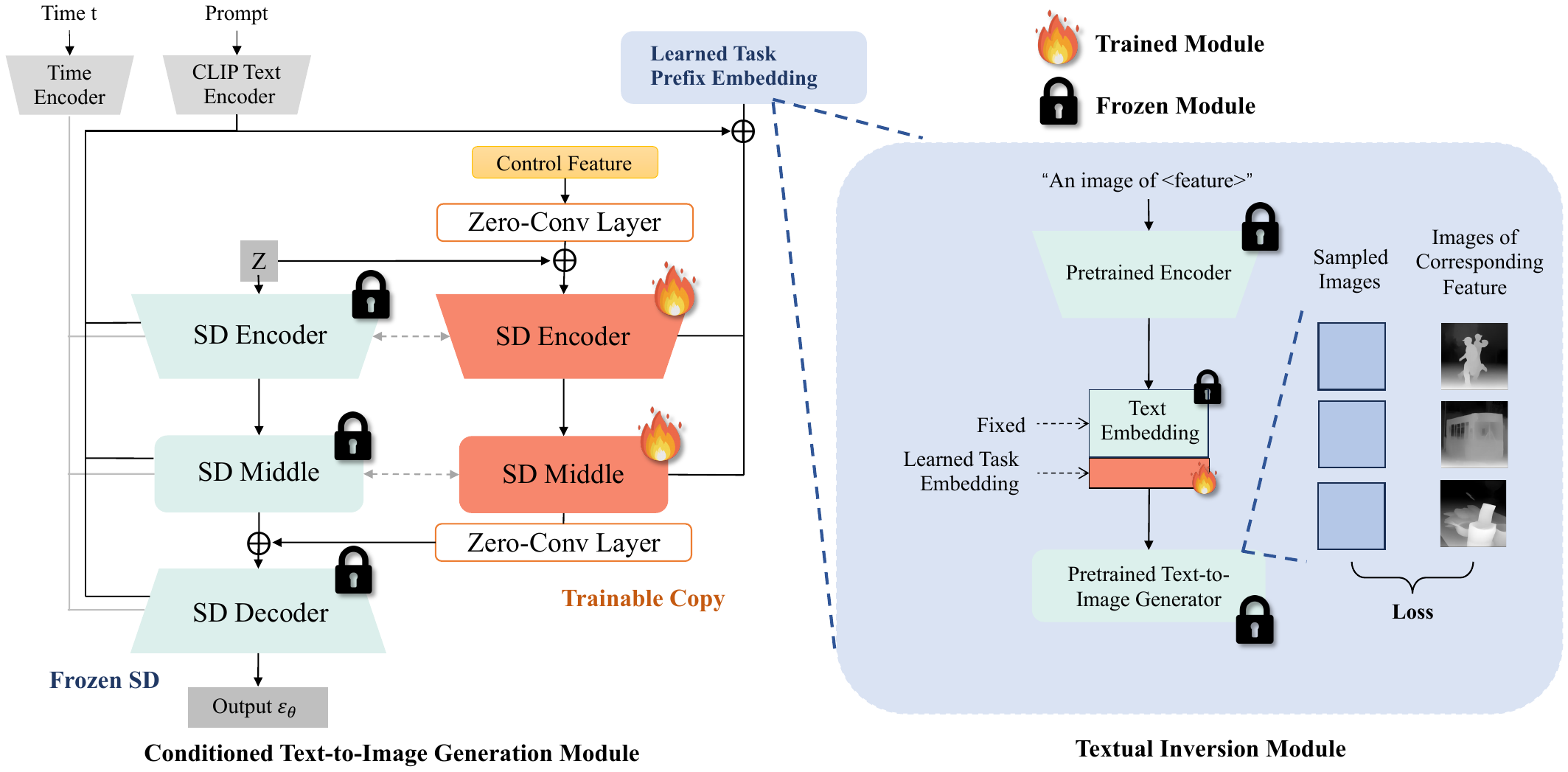}
    \caption{\textbf{An overview of our conditioned text-to-image generation pipeline.} Beginning with the original ControlNet structure \cite{zhang2023adding}, we utilize the textual inversion to learn task embeddings. Subsequently, we append the prefix \emph{use \textless feature\textgreater~as feature} to the prompt and feed the result into the trainable copy. The left side of the figure provides an overview of the conditioned text-to-image generation model, while the right side illustrates the process of learning the CLIP embedding for the new ``word" with textual inversion \cite{gal2022image}.}
    \label{fig:stage2}
\end{figure*}

\cref{fig:stage2} provides an overview of our conditioned text-to-image generation (stage 2) pipeline.

For different tasks, such as depth map or hed edge as an additional feature, we initially apply the textual inversion~\cite{gal2022image}, using 16 random images for each feature to learn the corresponding new ``words" (represented by forms such as \textless depth\textgreater~or \textless hed\textgreater). Subsequently, we add these new ``words" into the CLIP~\cite{NIPS2017_8a1d6947} embedding space so that when they are used in text prompts, the CLIP encoder can recognize their specific meanings.

After acquiring these new embeddings, we adapt the prompts for each (prompt, feature, image) triplet. For instance, if the feature for a given triplet is the depth map of the image and the original prompt is ``a motorcycle in front of a tree", the revised prompt would be ``Use \textless depth\textgreater~as a feature, a motorcycle in front of a tree". The modified triplets are fed into the trainable copy, while the corresponding original triplets are fed into the frozen part. Following this, the model is trained with a methodology similar to ControlNet, where the triplets are fed into the model undifferentiated, without separating them by features.

\cref{tab:model_comp} provides the comparison of the model size as well as the data scale when compared to other integrated models, including UniControl~\cite{qin2023unicontrol}, and Uni-ControlNet~\cite{zhao2023uni}, and our model demonstrates several advantages.

\begin{table}[h]
\centering
\scalebox{0.8}{
\begin{tabular}{l|c|c}
\toprule
\multicolumn{1}{l|}{\ } & \multicolumn{1}{c|}{\bfseries Extra Parameters} & \multicolumn{1}{c}{\bfseries Extra Data }\\
\midrule
Uni-ControlNet & 0 & $\times n$ \\
UniControl & 20M & None\\
\midrule
Ours & 0 & None\\
\bottomrule
\end{tabular}
}
\newline
\footnotesize{$^* n$ refers to the number of datasets we combine. In our work, $n=2$.}
\caption{Comparison of parameters and data scale between OmniControlNet and competing works. \emph{Extra Parameters} refers to the number of extra parameters compared to the original ControlNet, while \emph{Extra Data} refers to the increased amount of data during training. Uni-ControlNet needs to fill the blanks of the mixed datasets with black images, which will double the scale of the data.}
\label{tab:model_comp}
\end{table}

\textbf{When compared to UniControl}, our model, following the structure of ControlNet, requires no additional parameters. In contrast, UniControl incorporates an additional mixture-of-experts (MoE) module, resulting in a substantially larger model (20M more parameters than other models, including ControlNet, Uni-ControlNet, and our model). During training, an increase of 1 in batch size leads to a $\sim$3 Gigabytes increase in GPU memory usage.

\textbf{In contrast to Uni-ControlNet}, our model does not need to perform channel-wise concatenation of multiple additional features. In our configuration, different features originate from varying sets of images. Whereas for Uni-ControlNet, when an image provides a feature such as a depth map but lacks another (\eg animal pose), the corresponding channels for the animal pose are filled with zeros, yielding a larger data scale.

\subsection{Textual Inversion Module}

Textual Inversion \cite{gal2022image} is an approach for extracting and defining new concepts from a few example images, which is the inversion process of text-to-image generation. This method creates new ``words" or tokens in the embedding space of the text encoder within the text-to-image generation pipeline, such as Stable Diffusion \cite{rombach2022high}. Once established, these unique tokens can be integrated into textual prompts, allowing for precise control over the characteristics of the images produced.

We leverage Stable Diffusion as our base model. For the set of images provided, the prompt is set to $s=$ ``an image of \textless $w$\textgreater", while the embedded feature $v$ of the ``word" \textless $w$\textgreater is our target. For the frozen SD model, suppose $c$ is the encoded feature of $s$, then we can express $c=c(v)$, as $c$ is determined by $v$. Therefore, the optimization goal should be
$$
    v^*=\underset{v}{{\arg\min}}\ \mathbb{E}_{z\sim\varepsilon(x), \epsilon\sim\mathcal{N}(0, 1), c(v), t}||\epsilon-\epsilon_{\theta}(z_t, t, c(v))||_2^2.
$$
where $\theta$ is the weight of the UNet in the SD model and is frozen, and therefore we can directly simulate $v$ in this approach.

\subsection{The Whole Integrated Model}

Initially, we train the multi-task dense image prediction (stage 1) model, which can generate various features with a single model. Subsequently, the samples generated by the Stage 1 model serve as the training data for the conditioned text-to-image generation (stage 2) model. During inference, images are input into the Stage 1 model, whose output is then forwarded to the Stage 2 model for further processing. By utilizing this stage 1 model, we can directly sample different features from a single model without needing multiple expert models. Then, we can use these sampled features to generate images that share similar features with the original one but with specified semantic meanings. \cref{fig:wholepipeline} shows the structure of the whole pipeline.

\section{Experiments}
\label{sec:experiments}

\begin{table*}[htb]
    \centering
    \scalebox{0.9}{
\begin{tabular}{l|cccc|cccc}
    \hline
    \toprule
    \multicolumn{1}{c|}{\ } & \multicolumn{4}{c|}{\textbf{FID Score}} & \multicolumn{4}{c}{\textsc{CLIP}$_t$ \textbf{Similarity}}\\
    \cmidrule{2-5} \cmidrule{6-9}
    
    \multirow{-2}{*}{\textbf{Method}} & {\textbf{Depth} $\downarrow$}  & \textbf{HED} $\downarrow$ & \textbf{Scribble} $\downarrow$  & \textbf{Animal Pose} $\downarrow$  & \textbf{Depth} $\uparrow$ & \textbf{HED} $\uparrow$  & \textbf{Scribble} $\uparrow$  & \textbf{Animal Pose} $\uparrow$\\
    \midrule
    \textbf{Disunified Model} &   &   &   &   &  & &  &  \\
    \quad T2I-Adapter \cite{mou2023t2i} &  20.85 &  18.31 &  19.79 &  45.56 &  0.3099 & 0.3072 &  0.3094 &  0.3327 \\
    \quad ControlNet \cite{zhang2023adding} & 24.24 &  24.33 & 21.97 &  57.14 & 0.3076 & 0.2760 &  0.3091 &  0.3160 \\
    \midrule
    \textbf{Unified Stage 2} &   &   &   &   &  & &  &  \\
    \quad Uni-ControlNet \cite{zhao2023uni}& 33.71 & 28.56 & 30.24 & \textbf{47.71} & 0.3011 & \textbf{0.3072} & 0.3028 & \textbf{0.3321} \\
    \quad UniControl \cite{qin2023unicontrol}& 25.34 & \textbf{21.03} & 25.82 & 54.10 & 0.3020 & 0.3006 & \textbf{0.3043} & 0.3105 \\
    \cellcolorlightgray \quad Ours & \cellcolorlightgray \textbf{23.20} & \cellcolorlightgray 27.26 & \cellcolorlightgray \textbf{25.79} & \cellcolorlightgray 53.28 & \cellcolorlightgray \textbf{0.3055} & \cellcolorlightgray 0.2988 & \cellcolorlightgray 0.3002 &  \cellcolorlightgray 0.3292 \\
    \midrule
    \textbf{Unified Stage 1 + 2} &   &   &   &   &  & &  &  \\
    \cellcolorlightgray \quad Ours & \cellcolorlightgray 34.86 & \cellcolorlightgray 36.57 & \cellcolorlightgray  36.63 & \cellcolorlightgray  51.10 & \cellcolorlightgray  0.3024 & \cellcolorlightgray 0.2971 & \cellcolorlightgray 0.2971 & \cellcolorlightgray 0.3269  \\
    \bottomrule
    \hline
    \end{tabular}
    }
  \captionsetup{font=small}
  \caption{\textbf{Quantitative results} of our model, including single stage 2 (conditioned text-to-image generation) model and integrated stage 1 (multi-task dense image prediction) + integrated stage 2 (conditioned text-to-image generation) models. Although methods that utilize different models (T2I-Adapter and ControlNet) tend to perform better, our framework demonstrates competitive results among the integrated models. The numbers in bold indicate the best performance among the integrated methods. The \textbf{bold} numbers represent the best score among integrated methods.}
  \label{table:stage2}
\end{table*}

For our OmniControlNet and the competing works, we perform training and inference on 4 tasks, including Depth, HED, Scribble, and Animal Pose.

\subsection{Implementation Details}

\subsubsection{Datasets}

\noindent\textbf{Training.} The dataset for both multi-task dense image prediction (stage 1) and conditioned text-to-image generation (stage 2) training consists of 2 different parts. Features depth map, HED edge, and user scribble are from the first part, while the feature animal pose is from the second part. In the first part, we first use YOLOv5 \cite{redmon2016you} model to detect all the humans in the images from the Laion-5B~\cite{NEURIPS2022_a1859deb} dataset and choose the first 50,000 images that consist at most 1 human. We directly sample user scribbles from the images, employ an HED boundary detection model \cite{xie2015holistically} to generate HED edges, and use the Midas depth detector \cite{Ranftl2022} to produce depth maps. The captions of the images are taken from the origin Laion-5B dataset. In the second part, we utilize the AP-10K dataset \cite{yu2021ap} and use the MMPose \cite{mmpose2020} model to generate the animal poses of the animals. The captions are generated by the BLIP2 \cite{li2023blip2} model. In order to make the 2 parts contain approximately the same number of images, we duplicate each image in the second part 5 times.

\noindent\textbf{Sampling and Testing.} For the features depth map, HED edge, and user scribble, we utilize the validation split of the COCO2017~\cite{lin2014microsoft} dataset and obtain the corresponding feature in the same way as the training set. We use the first caption for each image in the dataset. For the animal pose, we utilize the APT-36K dataset \cite{yang2022aptk} and choose the first image from each frame as the dataset. We sample the animal poses the same way as the training set and use the BLIP2 \cite{li2023blip2} model to perform the image captioning.

\subsubsection{Training Details}

For our multi-task dense image prediction (stage 1) model, we assign distinct loss functions and associated weights for four different conditions. The depth map generation utilizes L1 loss, while binary cross-entropy loss is employed for the other three scenarios. The assigned loss weights for depth, HED edge, user scribble, and pose are 0.5, 1, 5, and 5, respectively. We resize all the images to 512$\times$512 and take a batch size 16. The model employs an SGD Optimizer with an initial learning rate of 1e-6, which subsequently decreases to 9e-7 following a polynomial decay pattern after 120k iterations. The entire training process takes about 20 hours on 8 NVIDIA RTX 3090 GPUs.

For the textual inversion module, each of the new ``word" of a corresponding feature is trained on 8 NVIDIA RTX 3090 GPUs for about 1 hour.

For our conditioned text-to-image generation (stage 2) model, the number of DDIM diffusion steps is set to 50. We adopt the AdamW optimizer and set the learning rate to 1e-5. We train the model on 8 NVIDIA RTX 3090 GPUs with batch size 2 for 50,000 iterations (4 epochs), which takes about 40 hours.

\subsubsection{Evaluation Metrics}

For our multi-task dense image prediction (stage 1) model, various metrics are adopted to evaluate different aspects of the model's performance. For depth estimation, the Root Mean Square Error (RMSE) is utilized. For edge detection, three distinct metrics are adopted: the fixed contour threshold (ODS), per-image best threshold (OIS), and average precision (AP). The ODS is a metric that evaluates edge detection performance by considering a fixed threshold value across all images, thereby providing a universal performance measure. On the other hand, OIS varies the threshold for each image to find the optimal threshold for that particular image, offering a more adaptive measure of performance. Lastly, AP is a commonly used metric in edge detection tasks. It computes the average precision value for recall values over the interval [0, 1].

For our conditioned text-to-image generation (stage 2) model and the integrated model, we adopt FID score \cite{radford2021learning} and CLIP$_t$ \cite{NIPS2017_8a1d6947} similarity score as our metrics. For the FID score, we utilize a widely used inception model to measure the similarity between synthesized and real images. For the CLIP$_t$ similarity score, for each pair of generated image and corresponding caption, we use ViT-B/32 \cite{dosovitskiy2020image} CLIP to encode them, and calculate the inner product of them as the CLIP$_t$ similarity score. We report the average of the inner products of all the image-caption pairs.

\subsection{Experiment Results}

\begin{figure*}[h!]
  \centering
    \includegraphics[width=1\textwidth]{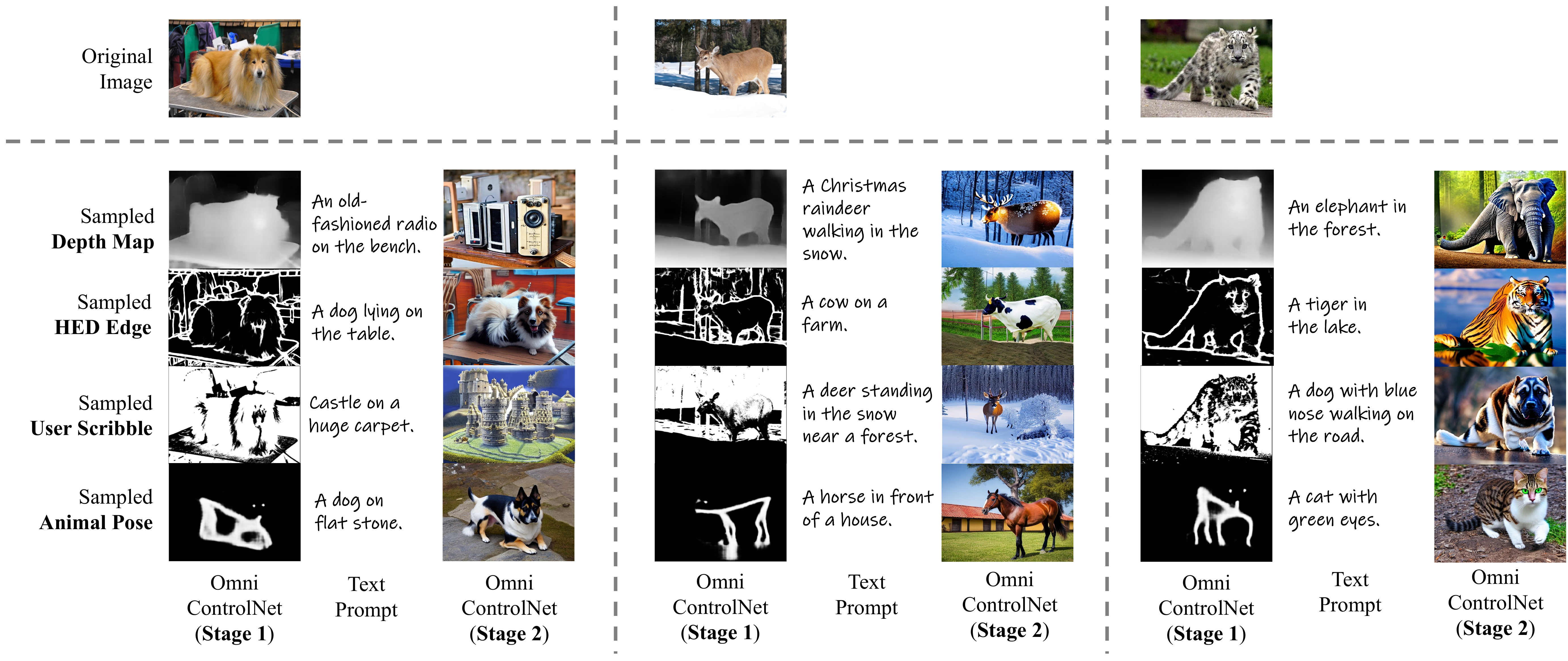}
    \caption{\small Features and images generated by our OmniControlNet model.}
    \label{fig:qualitative}
\end{figure*}

\cref{fig:teaser} and \cref{fig:qualitative} display the visual results for both the multi-task dense image prediction (stage 1), the conditioned text-to-image generation (stage 2), and the combined model. According to the figure, it is evident that the models from both stages and the combined one can generate high-quality results.

\vspace{0.5em}
\noindent\textbf{Stage 1: Integrated Dense Prediction.} To demonstrate the ability of our stage 1 model, we show the result on the depth benchmark NYUDv2~\cite{couprie2013nyudv2} and the HED benchmark BSDS500~\cite{gPb}. 

For depth estimation, we compare our result with DPT\textsubscript{hybrid}'s contemporary work, including DeepLabv3+ \cite{gur2019deeplabv3+}, RelativeDepth \cite{lee2019relativedepth}, ACAN \cite{chen2021acan}, ShapeNet \cite{ramamonjisoa2019sharpnet} and DPT\textsubscript{hybrid} \cite{Ranftl2021dpt}. As shown in \cref{tab:depth_detection}, our result outperforms all the models except for DPT\textsubscript{hybrid}. 

\begin{table}[!htp]
\centering
\scalebox{0.9}{
\begin{tabular}{l|c}
\toprule
\textbf{Method} & \textbf{RMSE $\downarrow$} \\
\midrule
DeepLabv3+ \cite{gur2019deeplabv3+} & 0.575 \\
RelativeDepth \cite{lee2019relativedepth} & 0.538 \\
ACAN \cite{chen2021acan} & 0.496 \\
ShapeNet \cite{ramamonjisoa2019sharpnet} & 0.496 \\
DPT\textsubscript{hybrid} \cite{Ranftl2021dpt} & 0.357 \\
\midrule
Ours & 0.472\\
\bottomrule
\end{tabular}
}
\caption{Depth performance of our multi-task dense image prediction (stage 1) model. Our model utilizes the output of DPT\textsubscript{hybrid} as the training data; therefore, it is acceptable for surpassing all other methods except for DPT\textsubscript{hybrid}.}
\label{tab:depth_detection}
\end{table}

For edge detection, we compare with classic methods including \cite{4767851canny, felzenszwalb2004efficient, arbelaez2010contour, ren2012discriminatively, lim2013sketch, isola2014crisp, dollar2014fast, hallman2015oriented, sironi2015projection, bertasius2015deepedge, hwang2015pixel, sironi2014multiscale, 7299024deepcontour, bertasius2015high, xie15hed}. As illustrated in \cref{tab:hed_detection}, our model surpasses all the models except for HED \cite{xie15hed}. 

\begin{table}[h]
\centering
\scalebox{0.85}{
\begin{tabular}{l|ccc}
\toprule
\textbf{Method} & \textbf{ODS $\uparrow$} & \textbf{OIS $\uparrow$} & \textbf{AP $\uparrow$} \\
\midrule
Canny \cite{4767851canny} &  0.600 & 0.640 & 0.580 \\
Felz-Hutt \cite{felzenszwalb2004efficient}  & 0.610 & 0.640 & 0.560 \\
gPb-owt-ucm \cite{arbelaez2010contour} & 0.726 & 0.757 & 0.696 \\
SCG \cite{ren2012discriminatively} & 0.739 & 0.758 & 0.773 \\
Sketch Tokens \cite{lim2013sketch} & 0.727 & 0.746 & 0.780 \\
PMI \cite{isola2014crisp} & 0.741 & 0.769 & 0.799 \\
SE \cite{dollar2014fast} & 0.746 & 0.767 & 0.803 \\
OEF \cite{hallman2015oriented} & 0.746 & 0.770 & 0.820 \\
MES \cite{sironi2015projection}  & 0.756 & 0.776 & 0.756 \\ 
DeepEdge \cite{bertasius2015deepedge} & 0.753 & 0.772 & 0.807 \\
CSCNN \cite{hwang2015pixel} & 0.756 & 0.775 & 0.798 \\
MSC \cite{sironi2014multiscale} & 0.756 & 0.776 & 0.787 \\
DeepContour \cite{shen2015deepcontour} & 0.757 & 0.776 & 0.800 \\
HFL \cite{bertasius2015high} & 0.767 & 0.788 & 0.795 \\
HED \cite{xie15hed} & 0.788 & 0.808 & 0.840 \\
\midrule
Ours& 0.761& 0.782 & 0.811\\
\bottomrule
\end{tabular}
}
\caption{HED performance of our multi-task dense image prediction (stage 1) model. For the three metrics, ODS, OIS, and AP, the larger the number, the better the performance. We can see that our method achieves competitive performance. }
\label{tab:hed_detection}
\end{table}

\vspace{0.5em}
\noindent\textbf{Stage 2: Integrated Conditioned Text-to-Image Generation.} We compare the quantitative results on the metrics FID score and CLIP$_t$ similarity score with other methods, including ControlNet \cite{zhang2023adding}, T2I-Adapter \cite{mou2023t2i}, Uni-ControlNet \cite{zhao2023uni} and UniControl \cite{qin2023unicontrol}. The latter two methods build an integrated pipeline that can use a single model to generate images with different additional features, while for the first two methods, a new model must be trained for each different additional feature.

\cref{table:stage2} presents the numerical results for the FID score and the CLIP$_t$~similarity score across various additional features and methods. Although methods that utilize different expert models for different features perform better, our method ranks among the best-performing methods within the category of integrated models.

\vspace{0.5em}
\noindent\textbf{Integrated Model Results.} In the integrated model, similar to the stage 2 model, we once again compare the quantitative results using metrics such as FID score and CLIP$_t$ similarity score with methods including T2I-Adapter~\cite{mou2023t2i}, ControlNet~\cite{zhang2023adding}, UniControl~\cite{qin2023unicontrol}, and Uni-ControlNet~\cite{zhao2023uni}. The quantitative results are presented in \cref{table:stage2}. It can be observed that although the overall performance of the integrated model is slightly inferior to methods directly utilizing features from multiple expert models, it still manages to generate images of promising quality.

\section{Ablation Studies}

To demonstrate the effectiveness of our model, OmniControlNet, and to reveal the impacts of certain structural designs, we conducted several ablation studies: \textbf{1)} Injecting learned task prefix embedding into different parts of the conditioned text-to-image generation module; \textbf{2)} Learning weights of the zero-convolution layers with an MLP while the model is trained with the learned task prefix embedding; and \textbf{3)} Comparing different encoding methods and the number of heads in the multi-head Feature Pyramid Network. For \textbf{1)} and \textbf{2)}, we report the results based on our unified stage 2 setting. For \textbf{3)}, we report the results based on our unified (stage 1 + stage 2) setting.

\subsection{Prefix Injection}
In our original framework, only the text prompts fed into the trainable copy of the SD model contain prefixes such as ``Use \textless depth\textgreater~as feature." In this ablation study, we added the prefix to both parts of the model. The results are shown in \cref{table:ablation_tivs}. We observe that adding the prefix only to the trainable part yields better results.

\begin{table}[h]
  \scalebox{0.71}{
  \begin{minipage}[ht]{\linewidth}
    \begin{tabular}{lcccc}
    \toprule
    \multicolumn{1}{l}{} & \multicolumn{4}{c}{\bfseries FID Scores}\\
    \cmidrule{2-5}
    \multicolumn{1}{l}{\bfseries Method} & {\bfseries Depth $\downarrow$} & {\bfseries HED $\downarrow$} & {\bfseries Scribble $\downarrow$} & {\bfseries Animal Pose $\downarrow$}\\
    \midrule
    Prefixes in both parts & 80.17 & 91.73 & 58.08 & 172.29\\
    \midrule
    OmniControlNet (Ours) & \textbf{23.20} & \textbf{27.26} & \textbf{25.79} & \textbf{53.28}\\
    \bottomrule
    \ \\
    \toprule
    \multicolumn{1}{l}{} & \multicolumn{4}{c}{\bfseries CLIP$_t$ Similarity Score}\\
    \cmidrule{2-5}
    \multicolumn{1}{l}{\bfseries Method} & {\bfseries Depth $\uparrow$} & {\bfseries HED $\uparrow$} & {\bfseries Scribble $\uparrow$} & {\bfseries Animal Pose $\uparrow$}\\
    \midrule
    Prefixes in both parts & 0.2321 & 0.2404 & 0.2676 & 0.1843\\
    \midrule
    OmniControlNet (Ours) & \textbf{0.3055} & \textbf{0.2988} & \textbf{0.3002} & \textbf{0.3292}\\
    \bottomrule
    \end{tabular}
  \end{minipage}%
  }
  \caption{Quantitative comparison of different prefix injection strategies. \emph{Prefixes in both parts} refers to adding a prefix to text prompts that are fed into both parts (frozen and trainable copy) of the model.}
  \vspace{-1em}
  \label{table:ablation_tivs}
\end{table}

\subsection{Learning Zero-Conv with MLP}

\begin{table}[htbp]
  \scalebox{0.71}{
  \begin{minipage}[t]{\linewidth}
    \begin{tabular}{lcccc}
    \toprule
    \multicolumn{5}{c}{\bfseries FID Scores}\\
    \midrule
    \multicolumn{1}{c}{\bfseries Method} & {\bfseries Depth $\downarrow$} & {\bfseries HED $\downarrow$} & {\bfseries Scribble $\downarrow$} & {\bfseries Animal Pose $\downarrow$}\\
    \midrule
    Learn weight by MLP & 32.06 & 32.17 & 32.04 & 72.21\\
    \midrule
    OmniControlNet (Ours) & \textbf{23.20} & \textbf{27.26} & \textbf{25.79} & \textbf{53.28}\\
    \bottomrule
    \\
    \toprule
    \multicolumn{5}{c}{\bfseries CLIP$_t$ Similarity Scores}\\
    \midrule
    \multicolumn{1}{c}{\bfseries Method} & {\bfseries Depth $\uparrow$} & {\bfseries HED $\uparrow$} & {\bfseries Scribble $\uparrow$} & {\bfseries Animal Pose $\uparrow$}\\
    \midrule
    Learn weight by MLP & \textbf{0.3102} & \textbf{0.3085} & \textbf{0.3101} & 0.3266\\
    \midrule
    OmniControlNet (Ours) & 0.3055 & 0.2988 & 0.3002 & \textbf{0.3292}\\
    \bottomrule
    \end{tabular}
  \end{minipage}%
  }
  \caption{Quantitative results of generating zero-conv weights via textual inversion embeddings. \emph{Learn weight by MLP} refers to the model using an MLP to learn the weight of the first zero-convolution.}
  \vspace{-0.5em}
  \label{table:ablation_mlp}
\end{table}

In our original framework, the zero-conv layers are initialed with zeros and updated during each training step by backpropagation, where multiple tasks share the same zero-conv weights. In the ablation study, we use an MLP to generate the weights of the first zero-conv layer from the textual inversion embedding of each task. The results are presented in \cref{table:ablation_mlp}. We observe that directly training the first convolution layer instead of using the MLP yields a better FID score, yet generating the weights dynamically via MLP produces an overall higher CLIP$_t$ score.

\subsection{Different Task Encoding and Number of Heads}

In our foundational framework, a multi-head Feature Pyramid Network (FPN) is employed to process multi-scale features, while one-hot encoded task embeddings are utilized for extracting target conditions. Our ablation study investigates the indispensability of the multi-head FPN and the efficacy of one-hot encoding. We implement two variations: one model with a single FPN head and another leveraging complex text embeddings generated by the CLIP \cite{patashnik2021styleclip} text encoder. The comparative results are detailed in \cref{tab:stage1_ablation}. Results show that integrating one-hot encoding with multiple FPN heads yields superior performance, demonstrating the effectiveness of our design.

\begin{table}[htb]
\centering
\scalebox{0.85}{
\begin{tabular}{l|ccc|c}
\toprule
\multicolumn{1}{l|}{\ } & \multicolumn{3}{c|}{\bfseries HED Edge} & \multicolumn{1}{c}{\bfseries Depth Map}\\
    \cmidrule{2-4} \cmidrule{5-5}
\multirow{-2}{*}{\textbf{Method}} & \textbf{ODS $\uparrow$} & \textbf{OIS $\uparrow$} & \textbf{AP $\uparrow$} & \textbf{RMSE $\downarrow$} \\
\midrule
Text Embedding & 0.600 & 0.640 & 0.580 & \ 0.558  \\
Single Head & 0.610 & 0.640 & 0.560 & \ 0.520  \\
\midrule
Ours & \textbf{0.761}& \textbf{0.782} & \textbf{0.811} & \ \textbf{0.472} \\
\bottomrule
\end{tabular}
}
\caption{Quantitative comparisons of different design choices of OmniControlNet. \emph{Text Embedding} refers to the model with CLIP \cite{patashnik2021styleclip} text encoded task embeddings. \emph{Single Head} delineates using a single-head Feature Pyramid Network (FPN).}
\vspace{-1em}
\label{tab:stage1_ablation}
\end{table}

\section{Conclusion and Limitations}

In this paper, we propose OmniControlNet, a streamlined approach that combines multiple external condition image generation processes into a cohesive one. This integration addresses the limitations of ControlNet's two-stage pipeline, which relies on external algorithms and has separate models for each input type. With OmniControlNet, we have a multitasking algorithm for generating conditions like edges, depth maps, and poses and an integrated image generation process guided by textual embedding. This results in a simpler, less redundant model capable of generating high-quality text-conditioned images.

\vspace{0.3em}
\noindent\textbf{Limitations.} \textbf{1)} When adding an additional task condition, it's required to train a new embedding for the task. \textbf{2)} With the integrated stage 1 model, the training complexity will increase, and image generation quality will decrease compared to using separate expert models as the stage 1 model.

\vspace{0.5em}
\noindent \textbf{Acknowledgement.} This work is supported by NSF Award IIS-2127544. We are grateful for the constructive feedbacks from Yifan Xu and Zheng Ding. 

\newpage

{
    \small
    \bibliographystyle{ieeenat_fullname}
    \bibliography{main}

\begin{thebibliography}{120}
\providecommand{\natexlab}[1]{#1}
\providecommand{\url}[1]{\texttt{#1}}
\expandafter\ifx\csname urlstyle\endcsname\relax
  \providecommand{\doi}[1]{doi: #1}\else
  \providecommand{\doi}{doi: \begingroup \urlstyle{rm}\Url}\fi

\bibitem[Alaluf et~al.(2021)Alaluf, Patashnik, and Cohen-Or]{alaluf2021only}
Yuval Alaluf, Or Patashnik, and Daniel Cohen-Or.
\newblock Only a matter of style: Age transformation using a style-based regression model.
\newblock \emph{ACM Transactions on Graphics (TOG)}, 40\penalty0 (4):\penalty0 1--12, 2021.

\bibitem[Arbelaez et~al.(2010)Arbelaez, Maire, Fowlkes, and Malik]{arbelaez2010contour}
Pablo Arbelaez, Michael Maire, Charless Fowlkes, and Jitendra Malik.
\newblock Contour detection and hierarchical image segmentation.
\newblock \emph{IEEE transactions on pattern analysis and machine intelligence}, 33\penalty0 (5):\penalty0 898--916, 2010.

\bibitem[Arbeláez et~al.(2011)Arbeláez, Maire, Fowlkes, and Malik]{gPb}
Pablo Arbeláez, Michael Maire, Charless Fowlkes, and Jitendra Malik.
\newblock Contour detection and hierarchical image segmentation.
\newblock \emph{IEEE Transactions on Pattern Analysis and Machine Intelligence}, 33\penalty0 (5):\penalty0 898--916, 2011.

\bibitem[Balaji et~al.(2022)Balaji, Nah, Huang, Vahdat, Song, Kreis, Aittala, Aila, Laine, Catanzaro, et~al.]{balaji2022ediffi}
Yogesh Balaji, Seungjun Nah, Xun Huang, Arash Vahdat, Jiaming Song, Karsten Kreis, Miika Aittala, Timo Aila, Samuli Laine, Bryan Catanzaro, et~al.
\newblock ediffi: Text-to-image diffusion models with an ensemble of expert denoisers.
\newblock \emph{arXiv preprint arXiv:2211.01324}, 2022.

\bibitem[Bertasius et~al.(2015{\natexlab{a}})Bertasius, Shi, and Torresani]{bertasius2015deepedge}
Gedas Bertasius, Jianbo Shi, and Lorenzo Torresani.
\newblock Deepedge: A multi-scale bifurcated deep network for top-down contour detection.
\newblock In \emph{Proceedings of the IEEE conference on computer vision and pattern recognition}, pages 4380--4389, 2015{\natexlab{a}}.

\bibitem[Bertasius et~al.(2015{\natexlab{b}})Bertasius, Shi, and Torresani]{bertasius2015high}
Gedas Bertasius, Jianbo Shi, and Lorenzo Torresani.
\newblock High-for-low and low-for-high: Efficient boundary detection from deep object features and its applications to high-level vision.
\newblock In \emph{Proceedings of the IEEE international conference on computer vision}, pages 504--512, 2015{\natexlab{b}}.

\bibitem[Birkl et~al.(2023)Birkl, Wofk, and M{\"u}ller]{birkl2023midas}
Reiner Birkl, Diana Wofk, and Matthias M{\"u}ller.
\newblock Midas v3.1 -- a model zoo for robust monocular relative depth estimation.
\newblock \emph{arXiv preprint arXiv:2307.14460}, 2023.

\bibitem[Brooks et~al.(2023)Brooks, Holynski, and Efros]{brooks2023instructpix2pix}
Tim Brooks, Aleksander Holynski, and Alexei~A Efros.
\newblock Instructpix2pix: Learning to follow image editing instructions.
\newblock In \emph{Proceedings of the IEEE/CVF Conference on Computer Vision and Pattern Recognition}, pages 18392--18402, 2023.

\bibitem[Canny(1986)]{4767851canny}
John Canny.
\newblock A computational approach to edge detection.
\newblock \emph{IEEE Transactions on Pattern Analysis and Machine Intelligence}, PAMI-8\penalty0 (6):\penalty0 679--698, 1986.

\bibitem[Cao et~al.(2017)Cao, Simon, Wei, and Sheikh]{cao2017realtime}
Zhe Cao, Tomas Simon, Shih-En Wei, and Yaser Sheikh.
\newblock Realtime multi-person 2d pose estimation using part affinity fields.
\newblock In \emph{CVPR}, 2017.

\bibitem[{Cao} et~al.(2019){Cao}, {Hidalgo Martinez}, {Simon}, {Wei}, and {Sheikh}]{8765346openpose}
Z. {Cao}, G. {Hidalgo Martinez}, T. {Simon}, S. {Wei}, and Y.~A. {Sheikh}.
\newblock Openpose: Realtime multi-person 2d pose estimation using part affinity fields.
\newblock \emph{IEEE Transactions on Pattern Analysis and Machine Intelligence}, 2019.

\bibitem[Chen et~al.(2021{\natexlab{a}})Chen, Wang, Guo, Xu, Deng, Liu, Ma, Xu, Xu, and Gao]{chen2021pre}
Hanting Chen, Yunhe Wang, Tianyu Guo, Chang Xu, Yiping Deng, Zhenhua Liu, Siwei Ma, Chunjing Xu, Chao Xu, and Wen Gao.
\newblock Pre-trained image processing transformer.
\newblock In \emph{Proceedings of the IEEE/CVF conference on computer vision and pattern recognition}, pages 12299--12310, 2021{\natexlab{a}}.

\bibitem[Chen et~al.(2023)Chen, Huang, Liu, Shen, Zhao, and Zhao]{chen2023anydoor}
Xi Chen, Lianghua Huang, Yu Liu, Yujun Shen, Deli Zhao, and Hengshuang Zhao.
\newblock Anydoor: Zero-shot object-level image customization.
\newblock \emph{arXiv preprint arXiv:2307.09481}, 2023.

\bibitem[Chen et~al.(2021{\natexlab{b}})Chen, Zhao, Hu, and Peng]{chen2021acan}
Yuru Chen, Haitao Zhao, Zhengwei Hu, and Jingchao Peng.
\newblock Attention-based context aggregation network for monocular depth estimation.
\newblock \emph{International Journal of Machine Learning and Cybernetics}, 12:\penalty0 1583--1596, 2021{\natexlab{b}}.

\bibitem[Choi et~al.(2018)Choi, Choi, Kim, Ha, Kim, and Choo]{choi2018stargan}
Yunjey Choi, Minje Choi, Munyoung Kim, Jung-Woo Ha, Sunghun Kim, and Jaegul Choo.
\newblock Stargan: Unified generative adversarial networks for multi-domain image-to-image translation.
\newblock In \emph{Proceedings of the IEEE conference on computer vision and pattern recognition}, pages 8789--8797, 2018.

\bibitem[Contributors(2020)]{mmpose2020}
MMPose Contributors.
\newblock Openmmlab pose estimation toolbox and benchmark.
\newblock \url{https://github.com/open-mmlab/mmpose}, 2020.

\bibitem[Couprie et~al.(2013)Couprie, Farabet, Najman, and LeCun]{couprie2013nyudv2}
Camille Couprie, Cl{\'e}ment Farabet, Laurent Najman, and Yann LeCun.
\newblock Indoor semantic segmentation using depth information.
\newblock \emph{arXiv preprint arXiv:1301.3572}, 2013.

\bibitem[Ding et~al.(2021)Ding, Yang, Hong, Zheng, Zhou, Yin, Lin, Zou, Shao, Yang, et~al.]{ding2021cogview}
Ming Ding, Zhuoyi Yang, Wenyi Hong, Wendi Zheng, Chang Zhou, Da Yin, Junyang Lin, Xu Zou, Zhou Shao, Hongxia Yang, et~al.
\newblock Cogview: Mastering text-to-image generation via transformers.
\newblock \emph{Advances in Neural Information Processing Systems}, 34:\penalty0 19822--19835, 2021.

\bibitem[Doll{\'a}r and Zitnick(2014)]{dollar2014fast}
Piotr Doll{\'a}r and C~Lawrence Zitnick.
\newblock Fast edge detection using structured forests.
\newblock \emph{IEEE transactions on pattern analysis and machine intelligence}, 37\penalty0 (8):\penalty0 1558--1570, 2014.

\bibitem[Dosovitskiy et~al.(2020)Dosovitskiy, Beyer, Kolesnikov, Weissenborn, Zhai, Unterthiner, Dehghani, Minderer, Heigold, Gelly, et~al.]{dosovitskiy2020image}
Alexey Dosovitskiy, Lucas Beyer, Alexander Kolesnikov, Dirk Weissenborn, Xiaohua Zhai, Thomas Unterthiner, Mostafa Dehghani, Matthias Minderer, Georg Heigold, Sylvain Gelly, et~al.
\newblock An image is worth 16x16 words: Transformers for image recognition at scale.
\newblock \emph{arXiv preprint arXiv:2010.11929}, 2020.

\bibitem[Esser et~al.(2021)Esser, Rombach, and Ommer]{esser2021taming}
Patrick Esser, Robin Rombach, and Bjorn Ommer.
\newblock Taming transformers for high-resolution image synthesis.
\newblock In \emph{Proceedings of the IEEE/CVF conference on computer vision and pattern recognition}, pages 12873--12883, 2021.

\bibitem[Felzenszwalb and Huttenlocher(2004)]{felzenszwalb2004efficient}
Pedro~F Felzenszwalb and Daniel~P Huttenlocher.
\newblock Efficient graph-based image segmentation.
\newblock \emph{International journal of computer vision}, 59:\penalty0 167--181, 2004.

\bibitem[Feng et~al.(2022)Feng, He, Fu, Jampani, Akula, Narayana, Basu, Wang, and Wang]{feng2022training}
Weixi Feng, Xuehai He, Tsu-Jui Fu, Varun Jampani, Arjun Akula, Pradyumna Narayana, Sugato Basu, Xin~Eric Wang, and William~Yang Wang.
\newblock Training-free structured diffusion guidance for compositional text-to-image synthesis.
\newblock \emph{arXiv preprint arXiv:2212.05032}, 2022.

\bibitem[Fu et~al.(2018)Fu, Gong, Wang, Batmanghelich, and Tao]{FuCVPR18-DORN}
Huan Fu, Mingming Gong, Chaohui Wang, Kayhan Batmanghelich, and Dacheng Tao.
\newblock {Deep Ordinal Regression Network for Monocular Depth Estimation}.
\newblock In \emph{{IEEE Conference on Computer Vision and Pattern Recognition (CVPR)}}, 2018.

\bibitem[Gal et~al.(2021)Gal, Patashnik, Maron, Chechik, and Cohen-Or]{gal2021stylegan}
Rinon Gal, Or Patashnik, Haggai Maron, Gal Chechik, and Daniel Cohen-Or.
\newblock Stylegan-nada: Clip-guided domain adaptation of image generators.
\newblock \emph{arXiv preprint arXiv:2108.00946}, 2021.

\bibitem[Gal et~al.(2022)Gal, Alaluf, Atzmon, Patashnik, Bermano, Chechik, and Cohen-Or]{gal2022image}
Rinon Gal, Yuval Alaluf, Yuval Atzmon, Or Patashnik, Amit~H Bermano, Gal Chechik, and Daniel Cohen-Or.
\newblock An image is worth one word: Personalizing text-to-image generation using textual inversion.
\newblock \emph{arXiv preprint arXiv:2208.01618}, 2022.

\bibitem[Goodfellow et~al.(2014)Goodfellow, Pouget-Abadie, Mirza, Xu, Warde-Farley, Ozair, Courville, and Bengio]{goodfellow2014generative}
Ian Goodfellow, Jean Pouget-Abadie, Mehdi Mirza, Bing Xu, David Warde-Farley, Sherjil Ozair, Aaron Courville, and Yoshua Bengio.
\newblock Generative adversarial nets.
\newblock \emph{Advances in neural information processing systems}, 2014.

\bibitem[Gu et~al.(2022)Gu, Ko, Go, Lee, Lee, and Shin]{gu2022towards}
Geonmo Gu, Byungsoo Ko, SeoungHyun Go, Sung-Hyun Lee, Jingeun Lee, and Minchul Shin.
\newblock Towards light-weight and real-time line segment detection.
\newblock In \emph{Proceedings of the AAAI Conference on Artificial Intelligence}, 2022.

\bibitem[Guo et~al.(2023)Guo, Yang, Rao, Wang, Qiao, Lin, and Dai]{guo2023animatediff}
Yuwei Guo, Ceyuan Yang, Anyi Rao, Yaohui Wang, Yu Qiao, Dahua Lin, and Bo Dai.
\newblock Animatediff: Animate your personalized text-to-image diffusion models without specific tuning.
\newblock \emph{arXiv preprint arXiv:2307.04725}, 2023.

\bibitem[Gur and Wolf(2019)]{gur2019deeplabv3+}
Shir Gur and Lior Wolf.
\newblock Single image depth estimation trained via depth from defocus cues.
\newblock In \emph{Proceedings of the IEEE/CVF Conference on Computer Vision and Pattern Recognition}, pages 7683--7692, 2019.

\bibitem[Hallman and Fowlkes(2015)]{hallman2015oriented}
Sam Hallman and Charless~C Fowlkes.
\newblock Oriented edge forests for boundary detection.
\newblock In \emph{Proceedings of the IEEE conference on computer vision and pattern recognition}, pages 1732--1740, 2015.

\bibitem[Hao et~al.(2023)Hao, Han, Zhao, and Wong]{hao2023vico}
Shaozhe Hao, Kai Han, Shihao Zhao, and Kwan-Yee~K Wong.
\newblock Vico: Detail-preserving visual condition for personalized text-to-image generation.
\newblock \emph{arXiv preprint arXiv:2306.00971}, 2023.

\bibitem[Hertz et~al.(2022)Hertz, Mokady, Tenenbaum, Aberman, Pritch, and Cohen-Or]{hertz2022prompt}
Amir Hertz, Ron Mokady, Jay Tenenbaum, Kfir Aberman, Yael Pritch, and Daniel Cohen-Or.
\newblock Prompt-to-prompt image editing with cross attention control.
\newblock \emph{arXiv preprint arXiv:2208.01626}, 2022.

\bibitem[Heusel et~al.(2017)Heusel, Ramsauer, Unterthiner, Nessler, and Hochreiter]{NIPS2017_8a1d6947}
Martin Heusel, Hubert Ramsauer, Thomas Unterthiner, Bernhard Nessler, and Sepp Hochreiter.
\newblock Gans trained by a two time-scale update rule converge to a local nash equilibrium.
\newblock In \emph{Advances in Neural Information Processing Systems}. Curran Associates, Inc., 2017.

\bibitem[Ho et~al.(2020)Ho, Jain, and Abbeel]{ho2020denoising}
Jonathan Ho, Ajay Jain, and Pieter Abbeel.
\newblock Denoising diffusion probabilistic models.
\newblock \emph{Advances in neural information processing systems}, 33:\penalty0 6840--6851, 2020.

\bibitem[Huan et~al.(2021)Huan, Xue, Zheng, He, Gong, and Xia]{huan2021unmixing}
Linxi Huan, Nan Xue, Xianwei Zheng, Wei He, Jianya Gong, and Gui-Song Xia.
\newblock Unmixing convolutional features for crisp edge detection.
\newblock \emph{IEEE Transactions on Pattern Analysis and Machine Intelligence}, 44\penalty0 (10):\penalty0 6602--6609, 2021.

\bibitem[Huang et~al.(2023{\natexlab{a}})Huang, Chen, Liu, Shen, Zhao, and Zhou]{huang2023composer}
Lianghua Huang, Di Chen, Yu Liu, Yujun Shen, Deli Zhao, and Jingren Zhou.
\newblock Composer: Creative and controllable image synthesis with composable conditions.
\newblock \emph{arXiv preprint arXiv:2302.09778}, 2023{\natexlab{a}}.

\bibitem[Huang et~al.(2023{\natexlab{b}})Huang, Tang, Dong, Lee, and Xu]{huang2023region}
Nisha Huang, Fan Tang, Weiming Dong, Tong-Yee Lee, and Changsheng Xu.
\newblock Region-aware diffusion for zero-shot text-driven image editing.
\newblock \emph{arXiv preprint arXiv:2302.11797}, 2023{\natexlab{b}}.

\bibitem[Hwang and Liu(2015)]{hwang2015pixel}
Jyh-Jing Hwang and Tyng-Luh Liu.
\newblock Pixel-wise deep learning for contour detection.
\newblock \emph{arXiv preprint arXiv:1504.01989}, 2015.

\bibitem[Isola et~al.(2014)Isola, Zoran, Krishnan, and Adelson]{isola2014crisp}
Phillip Isola, Daniel Zoran, Dilip Krishnan, and Edward~H Adelson.
\newblock Crisp boundary detection using pointwise mutual information.
\newblock In \emph{Computer Vision--ECCV 2014: 13th European Conference, Zurich, Switzerland, September 6-12, 2014, Proceedings, Part III 13}, pages 799--814. Springer, 2014.

\bibitem[Karras et~al.(2019)Karras, Laine, and Aila]{karras2019style}
Tero Karras, Samuli Laine, and Timo Aila.
\newblock A style-based generator architecture for generative adversarial networks.
\newblock In \emph{Proceedings of the IEEE/CVF conference on computer vision and pattern recognition}, pages 4401--4410, 2019.

\bibitem[Katzir et~al.(2022)Katzir, Perepelook, Lischinski, and Cohen-Or]{katzir2022multi}
Oren Katzir, Vicky Perepelook, Dani Lischinski, and Daniel Cohen-Or.
\newblock Multi-level latent space structuring for generative control.
\newblock \emph{arXiv preprint arXiv:2202.05910}, 2022.

\bibitem[Kendall et~al.(2015)Kendall, Grimes, and Cipolla]{kendall2015posenet}
Alex Kendall, Matthew Grimes, and Roberto Cipolla.
\newblock Posenet: A convolutional network for real-time 6-dof camera relocalization.
\newblock In \emph{Proceedings of the IEEE international conference on computer vision}, pages 2938--2946, 2015.

\bibitem[Kingma and Welling(2013)]{kingma2013auto}
Diederik~P Kingma and Max Welling.
\newblock Auto-encoding variational bayes.
\newblock 2013.

\bibitem[Kittler(1983)]{Kittler1983sobel}
Josef Kittler.
\newblock On the accuracy of the sobel edge detector.
\newblock \emph{Image Vis. Comput.}, 1:\penalty0 37--42, 1983.

\bibitem[Konishi et~al.(2003)Konishi, Yuille, Coughlan, and Zhu]{1159946statedge}
S. Konishi, A.L. Yuille, J.M. Coughlan, and Song~Chun Zhu.
\newblock Statistical edge detection: learning and evaluating edge cues.
\newblock \emph{IEEE Transactions on Pattern Analysis and Machine Intelligence}, 25\penalty0 (1):\penalty0 57--74, 2003.

\bibitem[Ladický et~al.(2014)Ladický, Shi, and Pollefeys]{6909413pull}
Lubor Ladický, Jianbo Shi, and Marc Pollefeys.
\newblock Pulling things out of perspective.
\newblock In \emph{2014 IEEE Conference on Computer Vision and Pattern Recognition}, pages 89--96, 2014.

\bibitem[Laina et~al.(2016)Laina, Rupprecht, Belagiannis, Tombari, and Navab]{laina2016deeper}
Iro Laina, Christian Rupprecht, Vasileios Belagiannis, Federico Tombari, and Nassir Navab.
\newblock Deeper depth prediction with fully convolutional residual networks.
\newblock In \emph{3D Vision (3DV), 2016 Fourth International Conference on}, pages 239--248. IEEE, 2016.

\bibitem[Lee and Kim(2019)]{lee2019relativedepth}
Jae-Han Lee and Chang-Su Kim.
\newblock Monocular depth estimation using relative depth maps.
\newblock In \emph{Proceedings of the IEEE/CVF Conference on Computer Vision and Pattern Recognition}, pages 9729--9738, 2019.

\bibitem[Li et~al.(2023{\natexlab{a}})Li, Li, Savarese, and Hoi]{li2023blip2}
Junnan Li, Dongxu Li, Silvio Savarese, and Steven Hoi.
\newblock Blip-2: Bootstrapping language-image pre-training with frozen image encoders and large language models, 2023{\natexlab{a}}.

\bibitem[Li et~al.(2023{\natexlab{b}})Li, Liu, Wu, Mu, Yang, Gao, Li, and Lee]{li2023gligen}
Yuheng Li, Haotian Liu, Qingyang Wu, Fangzhou Mu, Jianwei Yang, Jianfeng Gao, Chunyuan Li, and Yong~Jae Lee.
\newblock Gligen: Open-set grounded text-to-image generation.
\newblock In \emph{Proceedings of the IEEE/CVF Conference on Computer Vision and Pattern Recognition}, pages 22511--22521, 2023{\natexlab{b}}.

\bibitem[Lim et~al.(2013)Lim, Zitnick, and Doll{\'a}r]{lim2013sketch}
Joseph~J Lim, C~Lawrence Zitnick, and Piotr Doll{\'a}r.
\newblock Sketch tokens: A learned mid-level representation for contour and object detection.
\newblock In \emph{Proceedings of the IEEE conference on computer vision and pattern recognition}, pages 3158--3165, 2013.

\bibitem[Lin et~al.(2021)Lin, Men, Yang, Zhou, Ding, Zhang, Wang, Wang, Jiang, Jia, et~al.]{lin2021m6}
Junyang Lin, Rui Men, An Yang, Chang Zhou, Ming Ding, Yichang Zhang, Peng Wang, Ang Wang, Le Jiang, Xianyan Jia, et~al.
\newblock M6: A chinese multimodal pretrainer.
\newblock \emph{arXiv preprint arXiv:2103.00823}, 2021.

\bibitem[Lin et~al.(2014)Lin, Maire, Belongie, Hays, Perona, Ramanan, Doll{\'a}r, and Zitnick]{lin2014microsoft}
Tsung-Yi Lin, Michael Maire, Serge Belongie, James Hays, Pietro Perona, Deva Ramanan, Piotr Doll{\'a}r, and C~Lawrence Zitnick.
\newblock Microsoft coco: Common objects in context.
\newblock In \emph{Computer Vision--ECCV 2014: 13th European Conference, Zurich, Switzerland, September 6-12, 2014, Proceedings, Part V 13}, pages 740--755. Springer, 2014.

\bibitem[Lin et~al.(2017)Lin, Doll{\'a}r, Girshick, He, Hariharan, and Belongie]{lin2017fpn}
Tsung-Yi Lin, Piotr Doll{\'a}r, Ross Girshick, Kaiming He, Bharath Hariharan, and Serge Belongie.
\newblock Feature pyramid networks for object detection.
\newblock In \emph{Proceedings of the IEEE conference on computer vision and pattern recognition}, pages 2117--2125, 2017.

\bibitem[Liu et~al.(2015)Liu, Shen, Lin, and Reid]{liu2015learning}
Fayao Liu, Chunhua Shen, Guosheng Lin, and Ian Reid.
\newblock Learning depth from single monocular images using deep convolutional neural fields.
\newblock \emph{IEEE transactions on pattern analysis and machine intelligence}, 38\penalty0 (10):\penalty0 2024--2039, 2015.

\bibitem[Liu et~al.(2022)Liu, Li, Du, Torralba, and Tenenbaum]{liu2022compositional}
Nan Liu, Shuang Li, Yilun Du, Antonio Torralba, and Joshua~B Tenenbaum.
\newblock Compositional visual generation with composable diffusion models.
\newblock In \emph{European Conference on Computer Vision}, pages 423--439. Springer, 2022.

\bibitem[Liu et~al.(2021)Liu, Lin, Cao, Hu, Wei, Zhang, Lin, and Guo]{liu2021swin}
Ze Liu, Yutong Lin, Yue Cao, Han Hu, Yixuan Wei, Zheng Zhang, Stephen Lin, and Baining Guo.
\newblock Swin transformer: Hierarchical vision transformer using shifted windows.
\newblock In \emph{Proceedings of the IEEE/CVF international conference on computer vision}, pages 10012--10022, 2021.

\bibitem[Marr and Hildreth(1980)]{marr1980theory}
David Marr and Ellen Hildreth.
\newblock Theory of edge detection.
\newblock \emph{Proceedings of the Royal Society of London. Series B. Biological Sciences}, 207\penalty0 (1167):\penalty0 187--217, 1980.

\bibitem[Martin et~al.(2004)Martin, Fowlkes, and Malik]{Pb}
D.R. Martin, C.C. Fowlkes, and J. Malik.
\newblock Learning to detect natural image boundaries using local brightness, color, and texture cues.
\newblock \emph{IEEE Transactions on Pattern Analysis and Machine Intelligence}, 26\penalty0 (5):\penalty0 530--549, 2004.

\bibitem[Meng et~al.(2021)Meng, He, Song, Song, Wu, Zhu, and Ermon]{meng2021sdedit}
Chenlin Meng, Yutong He, Yang Song, Jiaming Song, Jiajun Wu, Jun-Yan Zhu, and Stefano Ermon.
\newblock Sdedit: Guided image synthesis and editing with stochastic differential equations.
\newblock \emph{arXiv preprint arXiv:2108.01073}, 2021.

\bibitem[Mou et~al.(2023)Mou, Wang, Xie, Zhang, Qi, Shan, and Qie]{mou2023t2i}
Chong Mou, Xintao Wang, Liangbin Xie, Jian Zhang, Zhongang Qi, Ying Shan, and Xiaohu Qie.
\newblock T2i-adapter: Learning adapters to dig out more controllable ability for text-to-image diffusion models.
\newblock \emph{arXiv preprint arXiv:2302.08453}, 2023.

\bibitem[Newell et~al.(2016)Newell, Yang, and Deng]{newell2016hourglass}
Alejandro Newell, Kaiyu Yang, and Jia Deng.
\newblock Stacked hourglass networks for human pose estimation.
\newblock In \emph{Computer Vision--ECCV 2016: 14th European Conference, Amsterdam, The Netherlands, October 11-14, 2016, Proceedings, Part VIII 14}, pages 483--499. Springer, 2016.

\bibitem[Park et~al.(2019)Park, Liu, Wang, and Zhu]{park2019semantic}
Taesung Park, Ming-Yu Liu, Ting-Chun Wang, and Jun-Yan Zhu.
\newblock Semantic image synthesis with spatially-adaptive normalization.
\newblock In \emph{Proceedings of the IEEE/CVF conference on computer vision and pattern recognition}, pages 2337--2346, 2019.

\bibitem[Patashnik et~al.(2021)Patashnik, Wu, Shechtman, Cohen-Or, and Lischinski]{patashnik2021styleclip}
Or Patashnik, Zongze Wu, Eli Shechtman, Daniel Cohen-Or, and Dani Lischinski.
\newblock Styleclip: Text-driven manipulation of stylegan imagery.
\newblock In \emph{Proceedings of the IEEE/CVF International Conference on Computer Vision}, pages 2085--2094, 2021.

\bibitem[Poole et~al.(2022)Poole, Jain, Barron, and Mildenhall]{poole2022dreamfusion}
Ben Poole, Ajay Jain, Jonathan~T Barron, and Ben Mildenhall.
\newblock Dreamfusion: Text-to-3d using 2d diffusion.
\newblock In \emph{International Conference on Learning Representations}, 2022.

\bibitem[Pu et~al.(2022)Pu, Huang, Liu, Guan, and Ling]{Pu_2022_CVPREDTER}
Mengyang Pu, Yaping Huang, Yuming Liu, Qingji Guan, and Haibin Ling.
\newblock Edter: Edge detection with transformer.
\newblock In \emph{Proceedings of the IEEE/CVF Conference on Computer Vision and Pattern Recognition (CVPR)}, pages 1402--1412, 2022.

\bibitem[Qin et~al.(2023)Qin, Zhang, Yu, Feng, Yang, Zhou, Wang, Niebles, Xiong, Savarese, et~al.]{qin2023unicontrol}
Can Qin, Shu Zhang, Ning Yu, Yihao Feng, Xinyi Yang, Yingbo Zhou, Huan Wang, Juan~Carlos Niebles, Caiming Xiong, Silvio Savarese, et~al.
\newblock Unicontrol: A unified diffusion model for controllable visual generation in the wild.
\newblock \emph{arXiv preprint arXiv:2305.11147}, 2023.

\bibitem[Radford et~al.(2021)Radford, Kim, Hallacy, Ramesh, Goh, Agarwal, Sastry, Askell, Mishkin, Clark, et~al.]{radford2021learning}
Alec Radford, Jong~Wook Kim, Chris Hallacy, Aditya Ramesh, Gabriel Goh, Sandhini Agarwal, Girish Sastry, Amanda Askell, Pamela Mishkin, Jack Clark, et~al.
\newblock Learning transferable visual models from natural language supervision.
\newblock In \emph{International conference on machine learning}, pages 8748--8763. PMLR, 2021.

\bibitem[Ramamonjisoa and Lepetit(2019)]{ramamonjisoa2019sharpnet}
Michael Ramamonjisoa and Vincent Lepetit.
\newblock Sharpnet: Fast and accurate recovery of occluding contours in monocular depth estimation.
\newblock In \emph{Proceedings of the IEEE/CVF International Conference on Computer Vision Workshops}, pages 0--0, 2019.

\bibitem[Ramesh et~al.(2021)Ramesh, Pavlov, Goh, Gray, Voss, Radford, Chen, and Sutskever]{ramesh2021zero}
Aditya Ramesh, Mikhail Pavlov, Gabriel Goh, Scott Gray, Chelsea Voss, Alec Radford, Mark Chen, and Ilya Sutskever.
\newblock Zero-shot text-to-image generation.
\newblock In \emph{International Conference on Machine Learning}, pages 8821--8831. PMLR, 2021.

\bibitem[Ramesh et~al.(2022)Ramesh, Dhariwal, Nichol, Chu, and Chen]{ramesh2022hierarchical}
Aditya Ramesh, Prafulla Dhariwal, Alex Nichol, Casey Chu, and Mark Chen.
\newblock Hierarchical text-conditional image generation with clip latents.
\newblock \emph{arXiv preprint arXiv:2204.06125}, 2022.

\bibitem[Ranftl et~al.(2020)Ranftl, Lasinger, Hafner, Schindler, and Koltun]{ranftl2020robust}
René Ranftl, Katrin Lasinger, David Hafner, Konrad Schindler, and Vladlen Koltun.
\newblock Towards robust monocular depth estimation: Mixing datasets for zero-shot cross-dataset transfer, 2020.

\bibitem[Ranftl et~al.(2021)Ranftl, Bochkovskiy, and Koltun]{Ranftl2021dpt}
Ren\'{e} Ranftl, Alexey Bochkovskiy, and Vladlen Koltun.
\newblock Vision transformers for dense prediction.
\newblock \emph{ICCV}, 2021.

\bibitem[Ranftl et~al.(2022{\natexlab{a}})Ranftl, Lasinger, Hafner, Schindler, and Koltun]{Ranftl2022}
Ren\'{e} Ranftl, Katrin Lasinger, David Hafner, Konrad Schindler, and Vladlen Koltun.
\newblock Towards robust monocular depth estimation: Mixing datasets for zero-shot cross-dataset transfer.
\newblock \emph{IEEE Transactions on Pattern Analysis and Machine Intelligence}, 44\penalty0 (3), 2022{\natexlab{a}}.

\bibitem[Ranftl et~al.(2022{\natexlab{b}})Ranftl, Lasinger, Hafner, Schindler, and Koltun]{Ranftl2022midas}
Ren\'{e} Ranftl, Katrin Lasinger, David Hafner, Konrad Schindler, and Vladlen Koltun.
\newblock Towards robust monocular depth estimation: Mixing datasets for zero-shot cross-dataset transfer.
\newblock \emph{IEEE Transactions on Pattern Analysis and Machine Intelligence}, 44\penalty0 (3), 2022{\natexlab{b}}.

\bibitem[Redmon et~al.(2016)Redmon, Divvala, Girshick, and Farhadi]{redmon2016you}
Joseph Redmon, Santosh Divvala, Ross Girshick, and Ali Farhadi.
\newblock You only look once: Unified, real-time object detection.
\newblock In \emph{Proceedings of the IEEE conference on computer vision and pattern recognition}, pages 779--788, 2016.

\bibitem[Reed et~al.(2016)Reed, Akata, Yan, Logeswaran, Schiele, and Lee]{reed2016generative}
Scott Reed, Zeynep Akata, Xinchen Yan, Lajanugen Logeswaran, Bernt Schiele, and Honglak Lee.
\newblock Generative adversarial text-to-image synthesis.
\newblock In \emph{Proceedings of The 33rd International Conference on Machine Learning}, 2016.

\bibitem[Ren and Bo(2012)]{ren2012discriminatively}
Xiaofeng Ren and Liefeng Bo.
\newblock Discriminatively trained sparse code gradients for contour detection.
\newblock In \emph{Proceedings of the 25th International Conference on Neural Information Processing Systems-Volume 1}, pages 584--592, 2012.

\bibitem[Richardson et~al.(2021)Richardson, Alaluf, Patashnik, Nitzan, Azar, Shapiro, and Cohen-Or]{richardson2021encoding}
Elad Richardson, Yuval Alaluf, Or Patashnik, Yotam Nitzan, Yaniv Azar, Stav Shapiro, and Daniel Cohen-Or.
\newblock Encoding in style: a stylegan encoder for image-to-image translation.
\newblock In \emph{Proceedings of the IEEE/CVF conference on computer vision and pattern recognition}, pages 2287--2296, 2021.

\bibitem[Rombach et~al.(2022)Rombach, Blattmann, Lorenz, Esser, and Ommer]{rombach2022high}
Robin Rombach, Andreas Blattmann, Dominik Lorenz, Patrick Esser, and Bj{\"o}rn Ommer.
\newblock High-resolution image synthesis with latent diffusion models.
\newblock In \emph{Proceedings of the IEEE/CVF conference on computer vision and pattern recognition}, pages 10684--10695, 2022.

\bibitem[Ronneberger et~al.(2015)Ronneberger, Fischer, and Brox]{ronneberger2015u}
Olaf Ronneberger, Philipp Fischer, and Thomas Brox.
\newblock U-net: Convolutional networks for biomedical image segmentation.
\newblock In \emph{Medical Image Computing and Computer-Assisted Intervention--MICCAI 2015: 18th International Conference, Munich, Germany, October 5-9, 2015, Proceedings, Part III 18}, pages 234--241. Springer, 2015.

\bibitem[Ruiz et~al.(2023)Ruiz, Li, Jampani, Pritch, Rubinstein, and Aberman]{ruiz2023dreambooth}
Nataniel Ruiz, Yuanzhen Li, Varun Jampani, Yael Pritch, Michael Rubinstein, and Kfir Aberman.
\newblock Dreambooth: Fine tuning text-to-image diffusion models for subject-driven generation.
\newblock In \emph{Proceedings of the IEEE/CVF Conference on Computer Vision and Pattern Recognition}, 2023.

\bibitem[Saharia et~al.(2022{\natexlab{a}})Saharia, Chan, Chang, Lee, Ho, Salimans, Fleet, and Norouzi]{saharia2022palette}
Chitwan Saharia, William Chan, Huiwen Chang, Chris Lee, Jonathan Ho, Tim Salimans, David Fleet, and Mohammad Norouzi.
\newblock Palette: Image-to-image diffusion models.
\newblock In \emph{ACM SIGGRAPH 2022 Conference Proceedings}, pages 1--10, 2022{\natexlab{a}}.

\bibitem[Saharia et~al.(2022{\natexlab{b}})Saharia, Chan, Saxena, Li, Whang, Denton, Ghasemipour, Gontijo~Lopes, Karagol~Ayan, Salimans, et~al.]{saharia2022photorealistic}
Chitwan Saharia, William Chan, Saurabh Saxena, Lala Li, Jay Whang, Emily~L Denton, Kamyar Ghasemipour, Raphael Gontijo~Lopes, Burcu Karagol~Ayan, Tim Salimans, et~al.
\newblock Photorealistic text-to-image diffusion models with deep language understanding.
\newblock \emph{Advances in Neural Information Processing Systems}, 35:\penalty0 36479--36494, 2022{\natexlab{b}}.

\bibitem[Saxena et~al.(2009)Saxena, Sun, and Ng]{4531745Make3D}
Ashutosh Saxena, Min Sun, and Andrew~Y. Ng.
\newblock Make3d: Learning 3d scene structure from a single still image.
\newblock \emph{IEEE Transactions on Pattern Analysis and Machine Intelligence}, 31\penalty0 (5):\penalty0 824--840, 2009.

\bibitem[Schuhmann et~al.(2022)Schuhmann, Beaumont, Vencu, Gordon, Wightman, Cherti, Coombes, Katta, Mullis, Wortsman, Schramowski, Kundurthy, Crowson, Schmidt, Kaczmarczyk, and Jitsev]{NEURIPS2022_a1859deb}
Christoph Schuhmann, Romain Beaumont, Richard Vencu, Cade Gordon, Ross Wightman, Mehdi Cherti, Theo Coombes, Aarush Katta, Clayton Mullis, Mitchell Wortsman, Patrick Schramowski, Srivatsa Kundurthy, Katherine Crowson, Ludwig Schmidt, Robert Kaczmarczyk, and Jenia Jitsev.
\newblock Laion-5b: An open large-scale dataset for training next generation image-text models.
\newblock In \emph{Advances in Neural Information Processing Systems}, pages 25278--25294. Curran Associates, Inc., 2022.

\bibitem[Shen et~al.(2015{\natexlab{a}})Shen, Wang, Wang, Bai, and Zhang]{7299024deepcontour}
Wei Shen, Xinggang Wang, Yan Wang, Xiang Bai, and Zhijiang Zhang.
\newblock Deepcontour: A deep convolutional feature learned by positive-sharing loss for contour detection.
\newblock In \emph{2015 IEEE Conference on Computer Vision and Pattern Recognition (CVPR)}, pages 3982--3991, 2015{\natexlab{a}}.

\bibitem[Shen et~al.(2015{\natexlab{b}})Shen, Wang, Wang, Bai, and Zhang]{shen2015deepcontour}
Wei Shen, Xinggang Wang, Yan Wang, Xiang Bai, and Zhijiang Zhang.
\newblock Deepcontour: A deep convolutional feature learned by positive-sharing loss for contour detection.
\newblock In \emph{Proceedings of the IEEE conference on computer vision and pattern recognition}, pages 3982--3991, 2015{\natexlab{b}}.

\bibitem[Singer et~al.(2022)Singer, Polyak, Hayes, Yin, An, Zhang, Hu, Yang, Ashual, Gafni, et~al.]{singer2022make}
Uriel Singer, Adam Polyak, Thomas Hayes, Xi Yin, Jie An, Songyang Zhang, Qiyuan Hu, Harry Yang, Oron Ashual, Oran Gafni, et~al.
\newblock Make-a-video: Text-to-video generation without text-video data.
\newblock \emph{arXiv preprint arXiv:2209.14792}, 2022.

\bibitem[Sironi et~al.(2014)Sironi, Lepetit, and Fua]{sironi2014multiscale}
Amos Sironi, Vincent Lepetit, and Pascal Fua.
\newblock Multiscale centerline detection by learning a scale-space distance transform.
\newblock In \emph{Proceedings of the IEEE Conference on Computer Vision and Pattern Recognition}, pages 2697--2704, 2014.

\bibitem[Sironi et~al.(2015)Sironi, Lepetit, and Fua]{sironi2015projection}
Amos Sironi, Vincent Lepetit, and Pascal Fua.
\newblock Projection onto the manifold of elongated structures for accurate extraction.
\newblock In \emph{Proceedings of the IEEE International Conference on Computer Vision}, pages 316--324, 2015.

\bibitem[Sohl-Dickstein et~al.(2015)Sohl-Dickstein, Weiss, Maheswaranathan, and Ganguli]{sohl2015deep}
Jascha Sohl-Dickstein, Eric Weiss, Niru Maheswaranathan, and Surya Ganguli.
\newblock Deep unsupervised learning using nonequilibrium thermodynamics.
\newblock In \emph{International conference on machine learning}, 2015.

\bibitem[Song et~al.(2020)Song, Meng, and Ermon]{song2020denoising}
Jiaming Song, Chenlin Meng, and Stefano Ermon.
\newblock Denoising diffusion implicit models.
\newblock \emph{arXiv preprint arXiv:2010.02502}, 2020.

\bibitem[Su et~al.(2022)Su, Song, Meng, and Ermon]{su2022dual}
Xuan Su, Jiaming Song, Chenlin Meng, and Stefano Ermon.
\newblock Dual diffusion implicit bridges for image-to-image translation.
\newblock \emph{arXiv preprint arXiv:2203.08382}, 2022.

\bibitem[Su et~al.(2021)Su, Liu, Yu, Hu, Liao, Tian, Pietik{\"a}inen, and Liu]{su2021pixel}
Zhuo Su, Wenzhe Liu, Zitong Yu, Dewen Hu, Qing Liao, Qi Tian, Matti Pietik{\"a}inen, and Li Liu.
\newblock Pixel difference networks for efficient edge detection.
\newblock In \emph{Proceedings of the IEEE/CVF international conference on computer vision}, pages 5117--5127, 2021.

\bibitem[Sun et~al.(2019)Sun, Xiao, Liu, and Wang]{sun2019deep}
Ke Sun, Bin Xiao, Dong Liu, and Jingdong Wang.
\newblock Deep high-resolution representation learning for human pose estimation.
\newblock In \emph{Proceedings of the IEEE/CVF conference on computer vision and pattern recognition}, pages 5693--5703, 2019.

\bibitem[Tu(2007)]{tu2007learning}
Zhuowen Tu.
\newblock Learning generative models via discriminative approaches.
\newblock In \emph{2007 IEEE Conference on Computer Vision and Pattern Recognition}, 2007.

\bibitem[Tumanyan et~al.(2023)Tumanyan, Geyer, Bagon, and Dekel]{tumanyan2023plug}
Narek Tumanyan, Michal Geyer, Shai Bagon, and Tali Dekel.
\newblock Plug-and-play diffusion features for text-driven image-to-image translation.
\newblock In \emph{Proceedings of the IEEE/CVF Conference on Computer Vision and Pattern Recognition}, pages 1921--1930, 2023.

\bibitem[Vasiljevic et~al.(2019)Vasiljevic, Kolkin, Zhang, Luo, Wang, Dai, Daniele, Mostajabi, Basart, Walter, and Shakhnarovich]{vasiljevic2019diode}
Igor Vasiljevic, Nick Kolkin, Shanyi Zhang, Ruotian Luo, Haochen Wang, Falcon~Z. Dai, Andrea~F. Daniele, Mohammadreza Mostajabi, Steven Basart, Matthew~R. Walter, and Gregory Shakhnarovich.
\newblock Diode: A dense indoor and outdoor depth dataset, 2019.

\bibitem[Vaswani et~al.(2017)Vaswani, Shazeer, Parmar, Uszkoreit, Jones, Gomez, Kaiser, and Polosukhin]{vaswani2017attention}
Ashish Vaswani, Noam Shazeer, Niki Parmar, Jakob Uszkoreit, Llion Jones, Aidan~N Gomez, {\L}ukasz Kaiser, and Illia Polosukhin.
\newblock Attention is all you need.
\newblock \emph{Advances in neural information processing systems}, 30, 2017.

\bibitem[Voynov et~al.(2023)Voynov, Aberman, and Cohen-Or]{voynov2023sketch}
Andrey Voynov, Kfir Aberman, and Daniel Cohen-Or.
\newblock Sketch-guided text-to-image diffusion models.
\newblock In \emph{ACM SIGGRAPH 2023 Conference Proceedings}, pages 1--11, 2023.

\bibitem[Wang et~al.(2022)Wang, Zhang, Zhang, Ouyang, Chen, Chen, and Wen]{wang2022pretraining}
Tengfei Wang, Ting Zhang, Bo Zhang, Hao Ouyang, Dong Chen, Qifeng Chen, and Fang Wen.
\newblock Pretraining is all you need for image-to-image translation.
\newblock \emph{arXiv preprint arXiv:2205.12952}, 2022.

\bibitem[Wang et~al.(2018)Wang, Liu, Zhu, Tao, Kautz, and Catanzaro]{wang2018high}
Ting-Chun Wang, Ming-Yu Liu, Jun-Yan Zhu, Andrew Tao, Jan Kautz, and Bryan Catanzaro.
\newblock High-resolution image synthesis and semantic manipulation with conditional gans.
\newblock In \emph{Proceedings of the IEEE conference on computer vision and pattern recognition}, pages 8798--8807, 2018.

\bibitem[Wang et~al.(2017)Wang, Zhao, and Huang]{wang2017deep}
Yupei Wang, Xin Zhao, and Kaiqi Huang.
\newblock Deep crisp boundaries.
\newblock In \emph{Proceedings of the IEEE conference on computer vision and pattern recognition}, pages 3892--3900, 2017.

\bibitem[Xiao et~al.(2018)Xiao, Wu, and Wei]{xiao2018simple}
Bin Xiao, Haiping Wu, and Yichen Wei.
\newblock Simple baselines for human pose estimation and tracking.
\newblock In \emph{Proceedings of the European conference on computer vision (ECCV)}, pages 466--481, 2018.

\bibitem["Xie and Tu(2015)]{xie15hed}
Saining "Xie and Zhuowen" Tu.
\newblock Holistically-nested edge detection.
\newblock In \emph{Proceedings of IEEE International Conference on Computer Vision}, 2015.

\bibitem[Xie and Tu(2015)]{xie2015holistically}
Saining Xie and Zhuowen Tu.
\newblock Holistically-nested edge detection.
\newblock In \emph{Proceedings of the IEEE international conference on computer vision}, pages 1395--1403, 2015.

\bibitem[Xu et~al.(2018{\natexlab{a}})Xu, Wang, Tang, Liu, Sebe, and Ricci]{xu2018structured}
Dan Xu, Wei Wang, Hao Tang, Hong Liu, Nicu Sebe, and Elisa Ricci.
\newblock Structured attention guided convolutional neural fields for monocular depth estimation.
\newblock In \emph{CVPR}, 2018{\natexlab{a}}.

\bibitem[Xu et~al.(2018{\natexlab{b}})Xu, Zhang, Huang, Zhang, Gan, Huang, and He]{xu2018attngan}
Tao Xu, Pengchuan Zhang, Qiuyuan Huang, Han Zhang, Zhe Gan, Xiaolei Huang, and Xiaodong He.
\newblock Attngan: Fine-grained text to image generation with attentional generative adversarial networks.
\newblock In \emph{Proceedings of the IEEE conference on computer vision and pattern recognition}, pages 1316--1324, 2018{\natexlab{b}}.

\bibitem[Yang et~al.(2017)Yang, Li, Ouyang, Li, and Wang]{yang2017learning}
Wei Yang, Shuang Li, Wanli Ouyang, Hongsheng Li, and Xiaogang Wang.
\newblock Learning feature pyramids for human pose estimation.
\newblock In \emph{proceedings of the IEEE international conference on computer vision}, pages 1281--1290, 2017.

\bibitem[Yang et~al.(2022)Yang, Yang, Xu, Zhang, Lan, and Tao]{yang2022aptk}
Yuxiang Yang, Junjie Yang, Yufei Xu, Jing Zhang, Long Lan, and Dacheng Tao.
\newblock {APT}-36k: A large-scale benchmark for animal pose estimation and tracking.
\newblock In \emph{Thirty-sixth Conference on Neural Information Processing Systems Datasets and Benchmarks Track}, 2022.

\bibitem[Yu et~al.(2021)Yu, Xu, Zhang, Zhao, Guan, and Tao]{yu2021ap}
Hang Yu, Yufei Xu, Jing Zhang, Wei Zhao, Ziyu Guan, and Dacheng Tao.
\newblock Ap-10k: A benchmark for animal pose estimation in the wild.
\newblock In \emph{Thirty-fifth Conference on Neural Information Processing Systems Datasets and Benchmarks Track (Round 2)}, 2021.

\bibitem[Yu et~al.(2022)Yu, Xu, Koh, Luong, Baid, Wang, Vasudevan, Ku, Yang, Ayan, et~al.]{yu2022scaling}
Jiahui Yu, Yuanzhong Xu, Jing~Yu Koh, Thang Luong, Gunjan Baid, Zirui Wang, Vijay Vasudevan, Alexander Ku, Yinfei Yang, Burcu~Karagol Ayan, et~al.
\newblock Scaling autoregressive models for content-rich text-to-image generation.
\newblock \emph{arXiv preprint arXiv:2206.10789}, 2022.

\bibitem[Zhang et~al.(2017)Zhang, Xu, Li, Zhang, Wang, Huang, and Metaxas]{zhang2017stackgan}
Han Zhang, Tao Xu, Hongsheng Li, Shaoting Zhang, Xiaogang Wang, Xiaolei Huang, and Dimitris~N Metaxas.
\newblock Stackgan: Text to photo-realistic image synthesis with stacked generative adversarial networks.
\newblock In \emph{Proceedings of the IEEE international conference on computer vision}, pages 5907--5915, 2017.

\bibitem[Zhang et~al.(2023)Zhang, Rao, and Agrawala]{zhang2023adding}
Lvmin Zhang, Anyi Rao, and Maneesh Agrawala.
\newblock Adding conditional control to text-to-image diffusion models.
\newblock In \emph{Proceedings of the IEEE/CVF International Conference on Computer Vision}, pages 3836--3847, 2023.

\bibitem[Zhao et~al.(2023)Zhao, Chen, Chen, Bao, Hao, Yuan, and Wong]{zhao2023uni}
Shihao Zhao, Dongdong Chen, Yen-Chun Chen, Jianmin Bao, Shaozhe Hao, Lu Yuan, and Kwan-Yee~K Wong.
\newblock Uni-controlnet: All-in-one control to text-to-image diffusion models.
\newblock \emph{Advances in Neural Information Processing Systems}, 2023.

\bibitem[Zhou et~al.(2019)Zhou, Zhao, Puig, Xiao, Fidler, Barriuso, and Torralba]{zhou2019semantic}
Bolei Zhou, Hang Zhao, Xavier Puig, Tete Xiao, Sanja Fidler, Adela Barriuso, and Antonio Torralba.
\newblock Semantic understanding of scenes through the ade20k dataset.
\newblock \emph{International Journal of Computer Vision}, 127\penalty0 (3):\penalty0 302--321, 2019.

\bibitem[Zhu et~al.(2017{\natexlab{a}})Zhu, Park, Isola, and Efros]{zhu2017unpaired}
Jun-Yan Zhu, Taesung Park, Phillip Isola, and Alexei~A Efros.
\newblock Unpaired image-to-image translation using cycle-consistent adversarial networks.
\newblock In \emph{Proceedings of the IEEE international conference on computer vision}, pages 2223--2232, 2017{\natexlab{a}}.

\bibitem[Zhu et~al.(2017{\natexlab{b}})Zhu, Zhang, Pathak, Darrell, Efros, Wang, and Shechtman]{zhu2017toward}
Jun-Yan Zhu, Richard Zhang, Deepak Pathak, Trevor Darrell, Alexei~A Efros, Oliver Wang, and Eli Shechtman.
\newblock Toward multimodal image-to-image translation.
\newblock \emph{Advances in neural information processing systems}, 30, 2017{\natexlab{b}}.

\end{thebibliography}
}

\end{document}